\documentclass[preprint,3p,10pt,sort&compress]{elsarticle}

\usepackage{subcaption}
\usepackage{siunitx}
\usepackage{placeins}
\usepackage{tikz}
\usetikzlibrary{positioning}
\usepackage{tikz-imagelabels}
\usepackage{pgfplots} 
\usepackage{pgfgantt}
\usepackage{pdflscape}
\pgfplotsset{compat=newest} 
\pgfplotsset{plot coordinates/math parser=false} 
\newlength\fwidth
\newlength\fheight

\usepackage{algorithm,algpseudocode}

\usepackage[colorlinks,citecolor=green,urlcolor=blue,bookmarks=false,hypertexnames=true]{hyperref} 

\usepackage{amsmath}
\usepackage{amssymb}
\usepackage{bm}
\usepackage{mathtools}
\usepackage{algorithm}
\usepackage{algpseudocode}
\usepackage{multirow}
\usepackage{upgreek}

\newcommand{\phiperp}{\mV_{\perp}}
\newcommand{\phipar}{\mV}

\newcommand{\mass}{m}
\newcommand{\rod}{l}

\newcommand{\va}{\mathbf{a}}
\newcommand{\vb}{\mathbf{b}}

\newcommand{\vq}{\mathbf{q}}
\newcommand{\vp}{\mathbf{p}}
\newcommand{\vm}{\mathbf{m}}
\newcommand{\vv}{\mathbf{v}}
\newcommand{\vw}{\mathbf{w}}
\newcommand{\vx}{\mathbf{x}}
\newcommand{\vy}{\mathbf{y}}
\newcommand{\vz}{\mathbf{z}}
\newcommand{\vtheta}{\boldsymbol{\uptheta}}

\newcommand{\vmu}{\boldsymbol{\upmu}}

\newcommand{\vzero}{\mathbf{0}}
\newcommand{\vone}{\mathbf{1}}

\newcommand{\rvh}{\bm{h}}

\newcommand{\rvv}{\bm{v}}
\newcommand{\rvw}{\bm{w}}
\newcommand{\rvx}{\bm{x}}
\newcommand{\rvy}{\bm{y}}
\newcommand{\rvz}{\bm{z}}
\newcommand{\rvtheta}{\boldsymbol{\theta}}
\newcommand{\rvxi}{\boldsymbol{\xi}}
\newcommand{\rveta}{\boldsymbol{\eta}}
\newcommand{\rvPsi}{\boldsymbol{\psi}}

\newcommand{\rtheta}{\boldsymbol{\theta}}

\newcommand{\rvxr}{\tilde{\rvx}}

\newcommand{\rvxpar}{\rvx_{\S}}
\newcommand{\rvxperp}{\rvx_{\S^{\perp}}}
\newcommand{\rvxpark}{\rvx_{k,\S}}
\newcommand{\rvxperpk}{\rvx_{k,\S^{\perp}}}

\newcommand{\mA}{\mathbf{A}}
\newcommand{\mC}{\mathbf{C}}

\newcommand{\mI}{\mathbf{I}}
\newcommand{\mK}{\mathbf{K}}
\newcommand{\mP}{\mathbf{P}}
\newcommand{\mQ}{\mathbf{Q}}

\newcommand{\mS}{\mathbf{S}}
\newcommand{\mU}{\mathbf{U}}
\newcommand{\mV}{\mathbf{V}}
\newcommand{\mX}{\mathbf{X}}

\newcommand{\mGamma}{\mathbf{\Gamma}}
\newcommand{\mSigma}{\mathbf{\Sigma}}
\newcommand{\mPhi}{\mathbf{\Phi}}

\newcommand{\Exp}[1]{\mathbb{E}\left[#1\right]}
\newcommand{\Cov}[2]{\mathbb{C}\text{ov}\left[#1,#2\right]}
\newcommand{\Var}[1]{\mathbb{V}\text{ar}\left[#1\right]}

\newcommand{\xcov}{\mC}
\newcommand{\pcovmult}{\mSigma_{\odot}}
\newcommand{\pcovadd}{\mSigma_{+}}
\newcommand{\mcovmult}{\mGamma_{\odot}}
\newcommand{\mcovadd}{\mGamma_{+}}

\newcommand{\reals}{\mathbb{R}}

\newcommand{\posint}{\mathbb{Z}_{+}}
\newcommand{\dimq}{d}
\newcommand{\dimr}{r}
\newcommand{\dimx}{n}
\newcommand{\dimy}{m}
\newcommand{\dimpar}{\ell}
\newcommand{\numy}{N}
\newcommand{\rmd}{{\rm d}}
\newcommand{\dt}{\Delta t}
\newcommand{\nz}{d}
\newcommand{\numsim}{\numy}

\renewcommand{\H}{\mathcal{H}}
\newcommand{\N}{\mathcal{N}}
\newcommand{\M}{\mathcal{M}}

\renewcommand{\S}{\mathcal{S}}
\newcommand{\T}{\mathcal{T}}
\newcommand{\U}{\mathcal{U}}

\newcommand{\probd}{\pi}
\newcommand{\like}{\mathcal{L}}

\newcommand{\yn}{\mathcal{Y}_{\numy}}
\newcommand{\ynr}{\tilde{\mathcal{Y}}_{\numy}}

\newcommand{\given}{|}
\newcommand{\yk}{\mathcal{Y}_k}
\newcommand{\ykk}{\mathcal{Y}_{k-1}}

\newcommand{\Inv}{\mathcal I}
\newcommand{\Invd}{I}

\newcommand{\y}{\mathbf y}

\newcommand{\q}{\mathbf q}
\newcommand{\qbar}{\bar{\mathbf q}}
\newcommand{\pbar}{\bar{\mathbf p}}
\newcommand{\Pp}{\mathbf P}
\newcommand{\qhat}{\tilde{\mathbf q}}
\newcommand{\phat}{\tilde{\mathbf p}}
\newcommand{\Qhat}{\tilde{\mathbf Q}}
\newcommand{\Phat}{\tilde{\mathbf P}}

\newcommand{\p}{\mathbf p}
\newcommand{\x}{\mathbf x}

\newcommand{\bzero}{\mathbf{0}}

\newcommand{\Fq}{\mathbf F_{\q}}
\newcommand{\Fp}{\mathbf F_{\p}}
\newcommand{\Fhatq}{\tilde{\mathbf F}_{\q}}
\newcommand{\Fhatp}{\tilde{\mathbf F}_{\p}}
\newcommand{\Rhatq}{\tilde{\mathbf R}_{\q}}
\newcommand{\Rhatp}{\tilde{\mathbf R}_{\p}}

\newcommand{\Dhatq}{\tilde{\mathbf D}_{\q}}
\newcommand{\Dhatp}{\tilde{\mathbf D}_{\p}}
\newcommand{\fq}{\mathbf f_{\q}}
\newcommand{\fp}{\mathbf f_{\p}}

\newcommand{\Jn}{\mathbf J_{2d}}
\newcommand{\Jr}{\mathbf J_{2r}}
\newcommand{\Vt}{\V^{+}}
\newcommand{\Real}{\mathbb{R}}
\newcommand{\V}{\mathbf V}

\newcommand{\vqr}{\tilde{\vq}}
\newcommand{\vpr}{\tilde{\vp}}
\newcommand{\vxr}{\tilde{\vx}}
\newcommand{\vyr}{\tilde{\vy}}
\newcommand{\vxpar}{\vx_{\S}}
\newcommand{\vxperp}{\vx_{\S^{\perp}}}
\newcommand{\vxpark}{\vx_{k,\S}}
\newcommand{\vxperpk}{\vx_{k,\S^{\perp}}}

\newcommand{\Hr}{\tilde{H}}

\newcommand{\diag}{\text{diag}}

\begin{document}
\begin{frontmatter}
\title{Bayesian identification of nonseparable Hamiltonians with multiplicative noise using deep learning and reduced-order modeling
}

\author[1]{Nicholas Galioto\corref{cor1}}
    \ead{ngalioto@umich.edu}

\author[2]{Harsh Sharma}
    \ead{hasharma@ucsd.edu}

\author[2]{Boris Kramer}
    \ead{bmkramer@ucsd.edu}    
  
\author[1]{Alex Arkady Gorodetsky}
    \ead{goroda@umich.edu}

  \affiliation[1]{organization={Department of Aerospace Engineering, University of Michigan},
    city={Ann Arbor},
    postcode={MI 48109},
    country={USA}}

\affiliation[2]{organization={Department of Mechanical and Aerospace Engineering, University of California San Diego},
    city={San Diego},
    postcode={CA 92161},
    country={USA}}

  \cortext[cor1]{Corresponding author}

\begin{highlights} 
    \item A Gaussian filter is derived for general additive and multiplicative noise models
    \item An algorithm is introduced for Bayesian estimation of high-dimensional Hamiltonians
    \item Probabilistic modeling is shown to improve a deep system ID method on noisy data
    \item Bayesian parameter estimation is successfully performed on a 64-dimensional system
\end{highlights}


\begin{abstract}
This paper presents a structure-preserving Bayesian approach for learning nonseparable Hamiltonian systems using stochastic dynamic models allowing for statistically-dependent, vector-valued additive and multiplicative measurement noise. The approach is comprised of three main facets. First, we derive a Gaussian filter for a statistically-dependent, vector-valued, additive and multiplicative noise model that is needed to evaluate the likelihood within the Bayesian posterior. Second, we develop a novel algorithm for cost-effective application of Bayesian system identification to high-dimensional systems. Third, we demonstrate how structure-preserving methods can be incorporated into the proposed framework, using nonseparable Hamiltonians as an illustrative system class. We assess the method's performance based on the forecasting accuracy of a model estimated from single-trajectory data. We compare the Bayesian method to a state-of-the-art machine learning method on a canonical nonseparable Hamiltonian model and a chaotic double pendulum model with small, noisy training datasets. The results show that using the Bayesian posterior as a training objective can yield upwards of 724 times improvement in Hamiltonian mean squared error using training data with up to 10\% multiplicative noise compared to a standard training objective. Lastly, we demonstrate the utility of the novel algorithm for parameter estimation of a 64-dimensional model of the spatially-discretized nonlinear Schr\"odinger equation with data corrupted by up to 20\% multiplicative noise.
\end{abstract}
\end{frontmatter}

\section{Introduction}

System identification (ID) plays a key role in many engineering and scientific frameworks such as model predictive control, system forecasting, and dynamical analysis. Creating a system ID algorithm includes careful selection of a class of candidate models and of an objective function to optimize. The success of system ID strongly depends on how efficiently this pair of model class and objective utilizes available information to guide estimation. Data from the system are most commonly used as sources of information, but prior knowledge on the system physics can also be considered within the estimation procedure.

Incorporating physical knowledge into system ID has been demonstrated through a variety of methodologies~\cite{karniadakis2021physics}. In one direction, physically-inconsistent models can be penalized through the addition of physics-based terms in the objective function. This approach is adopted by the widely-used physics-informed neural networks~\cite{raissi2019physics,linka2022bayesian} and has also been applied to various applications such as improving molecular dynamics simulation~\cite{zhang2007optimal} and lake temperature modeling~\cite{daw2020physics}. In another direction, physics are explicitly encoded into the model parameterization. This approach has led to various neural network architectures for learning conservative systems based on Hamiltonian~\cite{greydanus2019hamiltonian,chen2020symplectic,jin2020sympnets} and Lagrangian~\cite{saemundsson2020variational,cranmer2019lagrangian,lutter2018deep,roehrl2020modeling} mechanics. Other examples of this approach include preservation of symmetry groups in convolutional neural networks~\cite{mallat2016understanding} and enforcement of boundary conditions in boundary value problems~\cite{sheng2021pfnn}. Compared to traditional machine learning approaches, these methods have all shown significant improvements in estimation accuracy.

In addition to incorporating physical knowledge through models, data must be utilized to encourage consistency with the real world. Proper design of a learning objective is crucial to ensure information in the data is being extracted properly. The most common objectives take the form of a summation of vector norms of the differences between the data and estimated outputs, but alternative forms can be derived through probabilistic modeling. Notably, the modeling of model, measurement, and parameter uncertainties within a Bayesian framework in~\cite{galioto2020bayesian} led to an objective (the negative log posterior) that has been shown to yield more accurate estimates over many widely-used objectives~\cite{galioto2022likelihood}. To further improve information extraction capabilities, this method can be combined with structure-preserving parameterization techniques~\cite{galioto2020bayesian, sharma2022bayesian}. However, evaluation of this posterior relies on probabilistic filtering and is computationally challenging. Two challenges arise: (i) filtering can be costly if the noise models are non-additive, and (ii) filtering tends to not scale well with the state and measurement dimensions.

The challenge of non-additive noise arises when the sensor noise is dependent on the signal. The most common form of this is multiplicative noise. As an example, a distance-measuring sensor contains noise that increases roughly linearly with its distance from the target~\cite{hamill2001distance}. To address multiplicative noise, a Kalman filter was first derived in 1971 by~\cite{rajasekaran1971optimum}. Since then, a number of other filters have been developed for various multiplicative noise models such as Gaussian mixtures~\cite{liu2015optimal,wang2002robust}, non-stationary noise~\cite{chow1990new,zhang2007optimal,liu2015optimal}, and deterministic uncertainties~\cite{yang2002robust,kai2011robust}. Each of these filters, however, makes the assumption that the noise from each sensor is independent and identically distributed (i.i.d.). If sensors measure along the same axis, e.g., distance and velocity sensors or redundant sensors, their measurements, and therefore noises, will be correlated and not independent. And for sensors that have been manufactured differently, it is unlikely that their noise models will be identically distributed. Therefore, a more general filtering algorithm for multiplicative noises is needed.

The high-dimensional problem setting poses challenges for not only the Bayesian approach, but also many other system ID methods. Most nonlinear system ID algorithms require thousands of evaluations of a forward model (or data points) for training, which can quickly become prohibitive for large-scale dynamical models. This issue is often addressed through reduced-order modeling in which a low-dimensional approximation, known as a reduced-order model (ROM), of the high-dimensional dynamics is estimated. To identify the ROM, many approaches begin by estimating a low-dimensional subspace for the reduced-order state vector followed by system ID in the reduced-order space. In the time domain, common approaches for the subspace identification task include linear methods such as the proper orthogonal decomposition~\cite{berkooz1993proper} and the reduced basis method~\cite{rozza2008reduced,guo2019data}, and nonlinear methods such as autoencoders~\cite{kim2022fast,lee2020model} and polynomial manifolds~\cite{ShMuBuGeGlKr_symplecticMORmanifolds_2023,geelen2023operator}. These methods, however, are trained using full-field simulation data from full-order models (FOMs) that realistically are often incorrect, partially unknown, or uncertain. As a result, the accuracy of the ROMs is limited by the accuracy of the FOMs. To improve past this limit requires using experimental data collected directly from the system of interest. Since these data are often noisy, training with them introduces additional error into the subspace approximation. For a system ID method to handle this added error, careful modeling of uncertainty will be key.

In this work, we introduce methodologies to address the challenges of Bayesian system ID for multiplicative noise and high-dimensional systems. Then we demonstrate how these methodologies can be combined with structure-preserving methods, using nonseparable Hamiltonian systems as an example class of systems. Lastly, we apply the methodologies to estimate a dynamics model from single-trajectory data and evaluate the model's quality based on its forecasting accuracy. We choose nonseparable Hamiltonian systems for several reasons. Hamiltonian systems can demonstrate complex nonlinear behavior while possessing an underlying highly-structured geometry. These systems possess interesting physical properties that are important to preserve including conservation of the Hamiltonian, reversibility, and symplecticity. Nonseparable Hamiltonians, specifically, are of interest because they arise in diverse fields such as multibody dynamics and control in robotics~\cite{serra2019control}, the Kozai-Lidov mechanism in astrophysics~\cite{li2014chaos}, particle accelerators in physics~\cite{forest2006geometric}, 3D vortex dynamics in fluid mechanics~\cite{salmon1988hamiltonian}, and the nonlinear Schr\"odinger equation in quantum mechanics~\cite{colliander2010transfer}.

In a previous work~\cite{sharma2022bayesian}, we also considered Bayesian system ID of nonseparable Hamiltonian systems with multiplicative measurement noise. There are three major distinctions between that work and the present one. The first is that in the past work, we considered only a polynomial Hamiltonian with a polynomial model parameterization. The current work uses a more expressive deep neural network parameterization and considers a non-polynomial Hamiltonian example. Second, we previously used an additive noise model and trained on data with multiplicative noise to demonstrate robustness to model misspecification, but here we adapt the model and algorithm toward multiplicative noise. The third and most significant difference in this work is that we develop an original algorithm for efficient estimation of high-dimensional nonseparable Hamiltonian systems. These novel additions to the structure-preserving Bayesian learning framework of~\cite{sharma2022bayesian} are stated more specifically as follows:

\begin{itemize}
\item derivation of a Gaussian filter for a statistically-dependent, vector-valued, additive and multiplicative noise model and analysis of the added computational complexity compared to a filter for only additive noise in Section~\ref{sec:filter},
\item creation of a novel learning algorithm (Algorithm~\ref{alg:lowd_learning}) for estimation of high-dimensional nonseparable Hamiltonians in a reduced-dimensional space in Section~\ref{sec:rom},
\item numerical experimentation showing that the proposed likelihood-based objective outperforms the original objective of a state-of-the-art machine learning method when training with sparse data with multiplicative uniform noise with respect to both state and Hamiltonian error in Sections~\ref{sec:tao's example} and \ref{sec:double_pendulum}. These results include upwards of 724 times improvement in Hamiltonian mean squared error when training with data corrupted by up to 10\% multiplicative noise,
\item demonstration of the effectiveness of Algorithm~\ref{alg:lowd_learning} for parameter estimation on the 64-dimensional spatially-discretized nonlinear Schr\"{o}dinger equation within a reduced 8-dimensional space using data with 20\% multiplicative noise in Section~\ref{sec:nlse}.
\end{itemize}

The rest of this paper is structured as follows. Section~\ref{sec:background} provides notation and the probabilistic formulation of the system ID problem. Section~\ref{sec:design} gives a background discussion on algorithmic design choices for the Bayesian method. Section~\ref{sec:method} introduces a filter for a general additive and multiplicative noise model and reviews structure-preserving techniques for nonseparable Hamiltonian systems. Section~\ref{sec:rom} presents a novel algorithm for efficient and structure-preserving estimation of high-dimensional Hamiltonians. Section~\ref{sec:experiments} applies the proposed Bayesian algorithm to two low-dimensional (one polynomial and one non-polynomial) nonseparable Hamiltonian systems. Then, the novel algorithm for estimating high-dimensional Hamiltonians is applied to a spatial discretization of the nonlinear Schr\"odinger equation. Finally, Section~\ref{sec:conclusions} provides concluding remarks and future research directions.

\section{Problem statement}
\label{sec:background}
In this section, we define notation and formulate the system ID problem. 

\subsection{Notation}
The space of real numbers is denoted by $\reals$ and the set of positive integers by $\posint$. Vectors are written in lowercase, non-italic, bold font, e.g., $\vx$, and matrices in uppercase, non-italic, bold font, e.g., $\mA$. The transpose is denoted by the $^\top$ symbol.  If a vector varies in space and/or time, it is spatially indexed as $\vx^s$ and/or temporally indexed as $\vx_t$ for discrete space and time $s,t\in\posint$.

Let $(\Omega,\mathcal{F},\mathbb{P})$ be a probability triple where $\Omega$ is a sample space, $\mathcal{F}$ is a $\sigma$-algebra, and $\mathbb{P}$ is a probability measure. Random variables are denoted in lowercase, italic, bold font, e.g., $\rvz$, and their realizations are denoted as their non-italic counterparts, e.g., $\vz$. We assume that for a continuous random variable $\rvz$, the probability measure $\mathbb{P}(\rmd\rvz)$ admits a probability density function (pdf) $\probd(\vz)$. A $d$-dimensional uniform distribution with lower and upper bounds $\va,\vb\in\reals^{d}$ is denoted as $\U[\va,\vb]$. A normal distribution with mean $\vm$ and covariance $\mC$ is denoted as $\N(\vm,\mC)$. If $\rvz\sim\N(\vzero,\mC)$, then $\lvert\rvz\rvert$ follows a half-normal distribution denoted as $\text{half-}\N(\vzero,\mC)$.

The symbol $\odot$ represents element-wise multiplication. The $\odot$ operation is defined when the dimensions of the operands match or when the operands are a matrix and a vector with length equal to the number of matrix columns. In the latter case, $\odot$ multiplies the $i$th column of the matrix by the $i$th element of the vector. This operator has the useful property that $\rvz\odot\va\sim\N\left(\vm\odot\va, \mC\odot(\va\va^{\top})\right)$ for the Gaussian random vector $\rvz\sim\N(\vm,\mC)$ and a constant vector $\va$ of the same length.

\subsection{Probabilistic problem formulation}\label{sec:bayesid}
In this section, we describe a probabilistic problem formulation for Bayesian system ID for arbitrary noise models. We model the states $\rvx_k(\omega)\in\reals^{\dimx}$ and the outputs $\rvy_k(\omega)\in\reals^{\dimy}$ for $\omega\in\Omega$ as discrete-time stochastic processes indexed by $k\in\posint$. The dynamics are modeled as a hidden Markov model (HMM)
\begin{subequations}\label{eq:hmm_general}
\begin{align}
  \rvx_{k+1} &= \T(\rvx_k,\rvtheta,\omega), \\
  \rvy_{k}   &= \M(\rvx_k,\rvtheta,\omega),
\end{align}
\end{subequations}
where $\T:\reals^{\dimx}\times\reals^{\dimpar}\times\Omega\mapsto\reals^{\dimx}$ is the state-transition function and $\M:\reals^{\dimx}\times\reals^{\dimpar}\times\Omega\mapsto\reals^{\dimy}$ the measurement function. These functions are parameterized by the random variable $\rvtheta(\omega)\in\reals^{\dimpar}$, whose realizations we denote with $\vtheta$, and both operators are functions of $\omega\in\Omega$ to represent that their outputs are random variables. The system is therefore characterized by the sequences of transitional pdfs $\probd(\vx_{k+1}\given\vx_k,\vtheta)$ and conditional output pdfs $\probd(\vy_k\given\vx_k,\vtheta)$ induced by $\T$ and $\M$, respectively.

We seek to represent the posterior distribution $\probd(\vtheta\given\yn)$ characterizing the uncertainty in system parameters $\rvtheta$ given a collection of measurements $\yn\coloneqq(\vy_1,\ldots,\vy_{\numy})$. Bayes' rule expresses the posterior in a computable form via the likelihood $\like(\vtheta;\yn)\coloneqq\probd(\yn\given\vtheta)$, the prior $\probd(\vtheta)$, and a normalizing constant $\probd(\yn)$ known as the evidence according to
\begin{equation}
  \probd(\vtheta \given \yn) = \frac{\like(\vtheta;\yn)\probd(\vtheta)}{\probd(\yn)}.
  \label{eq:bayes}
\end{equation}
The HMM, however, has the additional collection of uncertain variables $\rvx_1,\ldots,\rvx_{\numy}$ about which we do not intend to make inferences. These uncertain states are marginalized out of the inference problem within the likelihood as $\int\probd(\yn,\vx_1,\ldots,\vx_{\numy}\given\vtheta)\rmd\vx_1,\ldots,\rmd\vx_{\numy}$. At first, this marginalization appears to require a costly $\dimx\numy$-dimensional integration. However, the recursive structure of the HMM can be exploited to break the high-dimensional integral into $\numy$ integrals of the more manageable dimension $\dimx$ by the decomposition
\begin{equation}
  \like(\vtheta;\yn) = \probd(\vy_1\given\vtheta)\prod_{k=2}^{\numy}\probd(\vy_k\given\vtheta,\ykk).
\end{equation}
Each term in this product can be efficiently computed using recursion, as shown in Algorithm~\ref{alg:marginal} from~\cite[Th. 12.3]{sarkka2013bayesian}.

\begin{algorithm}
  \caption{Recursive marginal likelihood evaluation~\cite{sarkka2013bayesian}}
  \label{alg:marginal}
  \begin{algorithmic}[1]{
    \Require $\probd(\vx_1|\vtheta)$, $\yn$
    \Ensure $\like(\vtheta;\yn)$
    \State Initialize $\probd(\vx_1|\mathcal{Y}_{0},\vtheta)\coloneqq\probd(\vx_1|\vtheta)$ and $\like(\vtheta;\mathcal{Y}_{0})\coloneqq1$    
    \For{$k=1,\ldots\numy$}
    \State Marginalize:
    $\displaystyle \probd(\vy_k | \mathcal{Y}_{k-1},\vtheta) \gets \int\probd(\vy_k|\vx_k,\vtheta)\probd(\vx_k|\mathcal{Y}_{k-1},\vtheta)\rmd\vx_k$ \vspace{1mm}\\
    \hspace{3.38cm}$\displaystyle \like(\vtheta;\mathcal{Y}_k) \gets \like(\vtheta;\mathcal{Y}_{k-1})\probd(\vy_k | \mathcal{Y}_{k-1},\vtheta)$
    \vspace{1mm}
    \If{$k < \numy$}
    \State {\raggedright Update: $\displaystyle\probd(\vx_{k}|\mathcal{Y}_{k},\vtheta) \gets \frac{\probd(\vy_k|\vx_k,\vtheta)}{\probd(\vy_k | \mathcal{Y}_{k-1},\vtheta)}\probd(\vx_k|\mathcal{Y}_{k-1},\vtheta)$}
    \State {\raggedright Predict: $\displaystyle\probd(\vx_{k+1}|\yk,\vtheta) \gets \int\probd(\vx_{k+1}|\vx_{k},\vtheta)\probd(\vx_{k}|\mathcal{Y}_{k},\vtheta)\rmd\vx_{k}$}
    \EndIf
    \EndFor}
  \end{algorithmic}
\end{algorithm}

\section{Algorithmic design choices}\label{sec:design}

While Algorithm~\ref{alg:marginal} is general enough to be applicable to any system that can be modeled in the general HMM form of Eq.~\eqref{eq:hmm_general}, this flexibility requires design choices addressing two primary challenges: (i) the computational expense of filtering and (ii) how to encode prior knowledge into the parameterizations of $\T$ and $\M$.

The first challenge of computational expense arises because the integrals in Algorithm~\ref{alg:marginal} do not, in general, admit closed form solutions. The user's choice in evaluation method will determine the accuracy of the marginal likelihood evaluation and overall computational complexity. The design choices to address the second challenge include selecting the coordinate system, system dimensions, and model fidelities within $\T$ and $\M$. These choices often involve a tradeoff between accuracy and computational expense. For the best accuracy and generalizability of the learned model, prior information on the system should inform the model parameterization as much as possible. We now describe these choices and our contributions in more detail.

\subsection{Sources of computational expense}\label{sec:integrals}
Integral evaluation methods within filtering can be divided into two classes: those that estimate the exact integral and those that estimate an approximation of the integral. Estimation of the exact integral is typically performed using sequential Monte Carlo algorithms such as particle filtering~\cite{andrieu2010particle}. The efficiency of this approach, however, is strongly dependent on the ability to draw uncorrelated samples from the appropriate distributions, which is, in general, nontrivial. When the efficiency of Monte Carlo sampling is prohibitive, approximations are used instead. The most common class of approximation methods for these integrals is Gaussian filtering. These methods approximate the marginal pdf $\like(\vtheta;\yk)$ as Gaussian, which requires tracking only the first two moments of all other pdfs in Algorithm~\ref{alg:marginal}. Tracking the mean and covariance of the prediction pdf $\probd(\vx_{k+1}\given\yk,\vtheta)$ and output pdf $\probd(\vy_k\given\ykk,\vtheta)$ can be achieved through linearization using Taylor series expansion or through Gaussian integration. These techniques are possible because the prediction and output pdfs are defined in terms of available functions, in this case $\T$ and $\M$. There is no such function, however, for the update pdf $\probd(\vx_k\given\yk,\vtheta)$. The mean and covariance of this pdf are instead approximated with the Kalman update, which delivers the minimum mean squared error (MMSE) estimate of these quantities.

\subsubsection{Process and measurement model forms}
If more is known about the forms of $\T$ and $\M$, linearization and Gaussian integration can sometimes be replaced by closed form solutions. To this end, it is useful to separate the dynamics and output functions into deterministic and stochastic components. Let $\Psi:\reals^{\dimx}\times\reals^{\dimpar}\mapsto\reals^{\dimx}$ and $h:\reals^{\dimx}\times\reals^{\dimpar}\mapsto\reals^{\dimy}$ represent deterministic dynamics and measurement functions, respectively. The stochastic components are usually further divided according to how they enter into the model. The two main approaches for modeling these components are as additive or multiplicative noise. If both of these types of noise are present, the model is written as
\begin{subequations}
  \label{eq:hmm}
  \begin{align}
    \T(\rvx_k,\rvtheta,\omega) &= \Psi(\rvx_k, \rvtheta)\odot\rvw_k(\omega) + \rvxi_k(\omega), \label{eq:dynamics}\\
    \M(\rvx_k,\rvtheta,\omega) &= h(\rvx_k, \rvtheta)\odot\rvv_k(\omega) + \rveta_k(\omega), \label{eq:observations}
  \end{align}
\end{subequations}
where $\rvxi_k(\omega),\rvw_k(\omega)\in\reals^{\dimx}$ and $\rveta_k(\omega),\rvv_k(\omega)\in\reals^{\dimy}$ are discrete-time stochastic processes.

The most common noise model is additive noise. One benefit of an additive noise model is that the first two moments of the noise terms can be estimated separately from those of the dynamics and observation functions, reducing complexity. Then the total mean and covariance for either $\T$ or $\M$ is simply the sum of these separate estimates. For linear systems, this allows for the means and covariances of $\probd(\vx_{k+1}\given\yk,\vtheta)$ and $\probd(\vy_k\given\ykk,\vtheta)$ to be evaluated analytically with the Kalman filter when the first two moments of the noise terms are known. Moreover, if the system is linear and the noise terms are Gaussian, then the prediction, output, and update pdfs in Algorithm~\ref{alg:marginal} are all Gaussian, and the Kalman filter is exact.

Although additive Gaussian noise is a suitable choice for many problems, recent works on learning Hamiltonians~\cite{wu2020structure,xiong2021nonseparable} have begun considering multiplicative uniform noise. As with any other noise model, Gaussian filtering can be used to approximate Algorithm~\ref{alg:marginal} for multiplicative uniform noise, but it is preferable to compute as much of each integral in closed form as possible for the sake of complexity. By using knowledge of the noise model, the general Gaussian filtering procedure can be adapted to replace a portion of the approximations with an exact and computationally efficient evaluation. In Section~\ref{sec:filter}, we introduce such a filtering procedure for multiplicative noise and analyze its computational expense.

\subsubsection{Dimensionality}

Although using Gaussian filtering for integral evaluation is significantly more efficient than Monte Carlo sampling, it tends to not scale well with the state and measurement dimensions. For example, the Kalman filter has a computational complexity of $\mathcal{O}(\numy(\dimx^3+\dimy^3))$, which can make optimization and sampling schemes infeasible for even moderate dimensions. Since in most real-world systems, $\dimx\gg\dimy$, the state dimension tends to be the limiting dimension. Therefore, one solution is to use reduced-order modeling to reduce the state dimension to $\dimr\ll\dimx$ and perform estimation in this $\dimr$-dimensional subspace. In Section~\ref{sec:rom}, we present a method for learning high-dimensional systems efficiently using Algorithm~\ref{alg:marginal} and reduced-order modeling. If this approach is used, it is critical that the ROM still preserve the geometric structure of the FOM. Structure-preservation can be achieved by encoding prior physics knowledge into the parameterization of the dynamics model $\T$. We discuss such an approach in the following section.

\subsection{Incorporation of prior knowledge}\label{sec:parameterization}
Here we consider the choice of model parameterization with a focus on embedding geometric structure into the model. The benefits of physics-informed parameterization are that it leads to models that generalize better beyond the training data and reduces the number of data required for training. There are two main ways to enforce the system physics within the dynamics model $\T$. First, the parameterization of $\T$ should be designed to only admit models whose dynamics possess the same physical structure of the system. Second, the model dynamics should be evaluated in such a way that the resulting flow preserves the structure of the dynamics. Since this work considers specialization of methods to nonseparable Hamiltonian systems, we briefly review aspects of the Hamiltonian structure that will inform the proposed parameterization method.

Finite-dimensional canonical Hamiltonian systems are defined by a scalar function $H$, known as the Hamiltonian, of the canonical position $\vq$ and momentum $\vp$, both in $\reals^{\dimq}$. For these systems, the state is defined as $\vx=\begin{bmatrix}\vq^{\top}&\vp^{\top}\end{bmatrix}^{\top}$ such that $\dimx=2\dimq$. The governing equations of the system, known as Hamilton's equations, are derived from this function
\begin{equation}\label{eq:derivative}
        \dot{\q}=\frac{\partial H(\q,\p)}{\partial \p}, \quad \quad \dot{\p}=-\frac{\partial H(\q,\p)}{\partial \q}.
\end{equation}
Many Hamiltonian systems of interest to engineers and scientists (e.g., \cite{serra2019control,forest2006geometric,salmon1988hamiltonian,li2014chaos,colliander2010transfer}) are not additively separable with respect to functions of the position and momentum. Such Hamiltonian systems are said to be \textit{nonseparable}. Unlike the separable Hamiltonian systems which can be written as $H(\q,\p)=T(\p) + U(\q)$ with a kinetic energy function $T(\p)$ and a potential energy function $U(\q)$, Eq.~\eqref{eq:derivative} cannot be further simplified for nonseparable Hamiltonians. A distinctive feature of Hamilton's equations is that they possess physically meaningful geometric properties that can be described in the form of symplecticity, invariants of motion, and energy conservation. We discuss how these properties can be embedded in the estimation procedure in Section~\ref{sec:integrator}.

\section{Methodology}\label{sec:method}

In this section, we present solutions to the problems of computational expense and prior knowledge incorporation. In Section~\ref{sec:filter}, we derive a Gaussian filter for a statistically-dependent, vector-valued, additive and multiplicative noise model, and we analyze its computational expense relative to a Gaussian filter for the more widely-used additive noise model. Then in Section~\ref{sec:integrator}, we add the additional capability of physics-informed estimation to the algorithm by embedding geometric structure within the dynamics propagator $\Psi$, and we tailor the approach specifically toward nonseparable Hamiltonian systems.

\subsection{Filtering with multiplicative noise}\label{sec:filter}

In this section, we extend the filter for multiplicative scalar noise from~\cite{rajasekaran1971optimum} to models with statistically-dependent, vector-valued noise. Consider the HMM with additive and multiplicative noise in Eq.~\eqref{eq:hmm}. Let the vector-valued multiplicative noise terms $\rvw_k$ and $\rvv_k$ each be i.i.d. with means $\bar{\vw}$ and $\bar{\vv}$ and covariances $\pcovmult$ and $\mcovmult$, respectively. Similarly, let the additive noise terms $\rvxi_k$ and $\rveta_k$ also be i.i.d. with zero means and covariances $\pcovadd$ and $\mcovadd$, respectively. Since we are interested in Gaussian filtering, the higher order moments of these noise terms are not needed. 

Recall from Section~\ref{sec:integrals} that the goal of Gaussian filtering within Algorithm~\ref{alg:marginal} is to compute (approximations of) the means and covariances of the distributions $\probd(\vx_k\given\ykk,\vtheta)$, $\probd(\vy_k\given\ykk,\vtheta)$, and $\probd(\vx_k\given\yk,\vtheta)$. Many of these evaluations require statistics of a function output that can be computed in closed form if the function is linear or approximated with Gaussian approximation if the function is nonlinear. To represent these function outputs, we denote $\Psi(\rvx_k,\rvtheta)$ as $\rvPsi_k$ and $h(\vy_k,\rvtheta)$ as $\rvh_k$. The following equations outline the filtering procedure with line numbers denoting where each group of equations are evaluated within Algorithm~\ref{alg:marginal}. The mean $\vm_k$ and covariance $\xcov_k$ of $\T(\rvx_{k-1},\rvtheta,\omega)$ with respect to $\probd(\vx_{k-1}\given\ykk,\vtheta)$ (line 6):
\begin{align}
  \label{eq:xmean}%
  \vm_k &= \Exp{\rvPsi_k}\odot\bar{\vw}, \\
  \label{eq:xvar}%
  \xcov_k &=  \Exp{\rvPsi_k\rvPsi_k^{\top}}\odot\pcovmult + \Var{\rvPsi_k}\odot(\bar{\vw}\bar{\vw}^{\top}) + \pcovadd.
\end{align} 
The mean $\vmu_k$ and covariance $\mS_k$ of $\M(\rvx_k,\rvtheta,\omega)$ with respect to $\probd(\vx_k\given\ykk,\vtheta)$ and the covariance $\mU_k$ between $\T(\rvx_{k-1},\rvtheta,\omega)$ and $\M(\rvx_k,\rvtheta,\omega)$ with respect to $\probd(\vx_k,\vx_{k-1}\given\ykk,\vtheta)$ (line 3): 
\begin{align}
  \label{eq:ymean}%
  \vmu_k &= \Exp{\rvh_k}\odot\bar{\vv}, \\
  \label{eq:yvar}%
  \mS_k &= \Exp{\rvh_k \rvh_k^{\top}}\odot\mcovmult + \Var{\rvh_k}\odot(\bar{\vv}\bar{\vv}^{\top})+\mcovadd, \\
  \label{eq:xycov}%
  \mU_k &= \Cov{\rvPsi_k}{\rvh_k}\odot\bar{\vv}_k.
\end{align}
The mean $\vm_k^+$ and covariance $\xcov_k^+$ of $\rvx_k$ with respect to $\probd(\vx_k\given\yk,\vtheta)$ (line 5):
\begin{align}
  \mK_k    &= \mU_k\mS_k^{-1}, \\
  \vm_{k}^+ &\approx \vm_k + \mK_k(\vy_k-\vmu_k), \\
  \xcov_k^+   &\approx \xcov_k - \mK_k\mU_k^{\top}.
\end{align}
These last three equations represent the linear MMSE estimator with gain matrix $\mK_k\in\reals^{\dimx\times\dimy}$. Therefore, $\vm_k^+$ and $\xcov_k^+$ are only approximations unless $\rvx_k$ and $\rvy_k$ are jointly Gaussian. Also notice that $\Exp{\rvPsi_k\rvPsi_k^{\top}}$ and $\Exp{\rvh_k\rvh_k^{\top}}$ are required when multiplicative noise is present. Again, these could be computed with linearization or Gaussian integration, but for computational efficiency, we assume that the function outputs are Gaussian. This assumption allows Eq.~\eqref{eq:xvar} to be computed in the following form:
\begin{equation}
    \xcov_k =  \Var{\rvPsi_k}\odot(\pcovmult+\bar{\vw}\bar{\vw}^{\top}) + (\Exp{\rvPsi_k}\Exp{\rvPsi_k}^{\top})\odot\pcovmult + \pcovadd.
\end{equation}
A similar form can be used to compute Eq.~\eqref{eq:yvar}.

In a past work~\cite{galioto2020bayesian}, we analyzed the computational complexity of the filtering procedure for additive noise models and found it to be on the order $\mathcal{O}(\numy(\dimx^3+\dimy^3))$. Here, we analyze the increase in computational complexity of this filtering procedure when multiplicative noise is present in addition to additive noise. Because we assumed that the noise is stationary, the outer products $\bar{\vw}\bar{\vw}^{\top}$ and $\bar{\vv}\bar{\vv}^{\top}$ need only be evaluated once. All other operations scale linearly with the number of data $\numy$. The added computational cost is summarized in Table~\ref{tab:complexity}. The order of the added expense is $\mathcal{O}(\numy(\dimx^2+\dimy^2))$, so the additional computation required when multiplicative noise is added to the model does not affect the order of complexity of the overall algorithm.

\begin{table}[ht]
\centering
  \caption{The increase in computational complexity of Gaussian filtering with both additive and multiplicative noise in the HMM~\eqref{eq:hmm} compared to Gaussian filtering with only additive noise present.}
  \label{tab:complexity}
  \begin{tabular}{|l|l|r|}
    \hline
    & Equation & Added flops \\ \hline
    \multirow{2}{*}{\textbf{Dynamics}}
    & Eq.~\eqref{eq:xmean} & $\numy\dimx$ \\ \cline{2-3}
    & Eq.~\eqref{eq:xvar} & $5\numy\dimx^2+\dimx^2$ \\ \hline
    \multirow{3}{*}{\textbf{Observations}}
    & Eq.~\eqref{eq:ymean} & $\numy\dimy$ \\ \cline{2-3}
    & Eq.~\eqref{eq:yvar} & $5\numy\dimy^2+\dimy^2$ \\ \cline{2-3}
    & Eq.~\eqref{eq:xycov} & $\numy\dimx\dimy$ \\ \hline
    \multicolumn{2}{|l|}{\textbf{Total:}} & \begin{tabular}[c]{@{}r@{}r}$\numy(5\dimx^2+5\dimy^2+\dimx\dimy$\\$+\dimx+\dimy)+\dimy^2+\dimx^2$\end{tabular} \\ \hline
  \end{tabular}
\end{table}

\subsection{Embedding symplectic structure}\label{sec:integrator} 

\begin{figure*}[ht]
  \centering
    \begin{tikzpicture}[
      squarednode/.style={rectangle, draw=black, fill=white, very thick, minimum height=2em},
      ]
      
    \node[squarednode,align=center] (node0) {Parameter value $\vtheta$};
    \node[squarednode,text width=7em,align=center] (node1) [below=2em of node0] {Parameterized Hamiltonian\\ $H(\vq,\vp,\vtheta)$};
    \node[squarednode,text width=11em,align=center] (node2) [right=2em of node1] {Hamilton's equations~\eqref{eq:derivative}\\$\dot{\vq}(\vtheta),\dot{\vp}(\vtheta)$};
    \node[squarednode,text width=7em,align=center] (node3) [right=2em of node2] {Symplectic integrator~\eqref{eq:exp_symp}};
    \node[squarednode,text width=10em,align=center] (node4) [right=2em of node3] {Structure-preserving propagator $\Psi(\vq,\vp,\vtheta)$};
    \node[squarednode,align=center] (node5) [above=2em of node4] {Likelihood evaluation $\probd(\vtheta\given\yn)$};

    \draw[->, ultra thick](node0.south) -- (node1.north);
    \draw[->, ultra thick] (node1.east) -- (node2.west);
    \draw[->, ultra thick] (node2.east) -- (node3.west);
    \draw[->, ultra thick] (node3.east) -- (node4.west);
    \draw[->, ultra thick](node4.north) -- (node5.south);
    \end{tikzpicture}
  \caption{Schematic showing how Hamiltonian structure is preserved when evaluating the likelihood.}
  \label{fig:propagator}
\end{figure*}
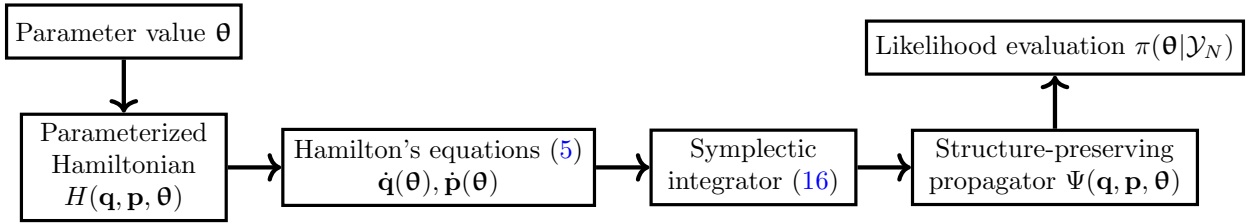

Next we describe a parameterization strategy for embedding symplectic structure within the learning process that is shown graphically in Fig.~\ref{fig:propagator}. The main idea behind this strategy is that rather than parameterizing the time derivatives or the propagator $\Psi$ directly, we parameterize the Hamiltonian $H(\vq,\vp,\rvtheta)$. From this Hamiltonian, the time derivatives are derived from Hamilton's equations and then integrated with a symplectic integrator. This process ensures that the estimated model will be Hamiltonian and also that its flow will be symplectic. This approach has shown utility in various other works~\cite{galioto2020symplectic,sharma2022bayesian,chen2020symplectic,david2023symplectic}.

From the parameterized Hamiltonian, the continuous-time dynamics are given by Hamilton's equations~\eqref{eq:derivative}. However, the HMM~\eqref{eq:hmm} that we are using is formulated in discrete time. To resolve this discrepancy, we require a propagator $\Psi$ that can map the state forward in time using the estimated time derivatives without violating the physical properties mentioned in Section~\ref{sec:parameterization}. For this purpose, we utilize a recently-developed explicit symplectic integrator for nonseparable Hamiltonian systems~\cite{tao2016explicit} that we refer to as ``Tao's integrator'' throughout the paper. We briefly describe the integration method here.

Consider an arbitrary Hamiltonian $H(\q,\p)$ and fictitious position and momentum vectors $\qbar$ and $\pbar$ corresponding to $\q$ and $\p$. Next, define the augmented Hamiltonian
\begin{equation}\label{eq:aug}
    \bar{H}(\q,\p,\qbar,\pbar):=\underbrace{H(\q,\pbar)}_{H_a} + \underbrace{H(\qbar,\p)}_{H_b} + \lambda\cdot \underbrace{ \left( \Vert \q-\qbar \Vert_2^2/2 + \Vert \p-\pbar \Vert_2^2/2 \right)}_{H_c},
\end{equation}
where $H_a:=H(\q,\pbar)$ and $H_b:=H(\qbar,\p)$ correspond to two copies of the original nonseparable Hamiltonian system with mixed-up positions and momenta; $H_c$ is an artificial constraint; and $\lambda$ is a constant that controls the binding of the two copies. Unlike the original Hamiltonian $H$, the extended Hamiltonian $\bar{H}$ is amenable to explicit symplectic integration.

From $\bar{H}$, a propagator $\Psi$ is derived using a second-order explicit symplectic method based on Strang splitting
\begin{equation}
\Psi:=\psi^{\Delta t/2}_{H_a} \circ \psi^{\Delta t/2}_{H_b}\circ \psi^{\Delta t}_{\lambda H_c} \circ \psi^{\Delta t/2}_{H_b}\circ \psi^{\Delta t/2}_{H_a},
\label{eq:exp_symp}
\end{equation} 
where $\psi^{\Delta t}_{H_a}, \psi^{\Delta t}_{H_b},$ and $\psi^{\Delta t}_{\lambda H_c}$ are the time-$\dt$ flow of $H_a,H_b,$ and $\lambda H_c$. Each flow is evaluated with explicit symplectic Euler substeps, so the result of the composition is an explicit and symplectic propagator as desired. 

\section{Structure-preserving dimension reduction for high-dimensional systems}\label{sec:rom}
\begin{figure*}[ht]
  \centering
  \begin{tikzpicture}[
      squarednode/.style={rectangle, draw=black, fill=white, very thick, minimum height=2em},
      hamiltonnode/.style={rectangle, draw=red, fill=white, very thick, minimum height=2em},
    ]

    \node[squarednode,align=center] (node0) {Parameter value $\vtheta_{\text{nl}}$};
    \node[squarednode,align=center] (ham0) [right=2em of node0] {H-OpInf~\cite{sharma2022hamiltonian} yields $\vtheta_{\text{quad}}$};
    \node[squarednode,text width=10em,align=center] (node1) [below right = 2em and -10em of node0] {Parameterized low-dimensional Hamiltonian~\eqref{eq:Hr_params} $\tilde{H}(\vqr,\vpr,\vtheta)$};
    \node[squarednode,text width=11em,align=center] (node2) [right=2em of node1] {Low-dimensional Hamilton's equations~\eqref{eq:dt_param}\\$\dot{\vqr}(\vtheta),\dot{\vpr}(\vtheta)$};
    \node[squarednode,text width=7em,align=center] (node3) [right=2em of node2] {Symplectic integrator~\eqref{eq:exp_symp}};
    \node[squarednode,text width=9em,align=center] (node4) [right=2em of node3] {Structure-preserving low-dimensional propagator $\tilde{\Psi}(\vqr,\vpr,\vtheta)$};
    \node[squarednode,align=center,text width=6em] (node5) [above=2em of node4] {Likelihood evaluation $\probd(\vtheta_{\text{nl}}\given\ynr)$};
    \node[squarednode,align=center,text width=9em,align=center] (node6) [left=1.5em of node5] {Low-dimensional observation model $\tilde{h}(\vqr,\vpr,\vtheta)$};

    \draw[->, ultra thick] (node0) -- (node1) [midway, above];
    \draw[->, ultra thick] (node0.east) -- (ham0.west);
    \draw[->, ultra thick] (ham0) -- (node1) [midway, above];
    \draw[->, ultra thick] (node1.east) -- (node2.west);
    \draw[->, ultra thick] (node2.east) -- (node3.west);
    \draw[->, ultra thick] (node3.east) -- (node4.west);
    \draw[->, ultra thick] (node4.north) -- (node5.south);
    \draw[->, ultra thick] (node6.east) -- (node5.west);
  \end{tikzpicture}
  \caption{Schematic showing how Hamiltonian structure is preserved in reduced dimensions when evaluating the likelihood. In this setting, the parameter vector is partitioned into quadratic $\theta_{\text{quad}}$ and nonlinear components $\theta_{\text{nl}}$, where $\theta_{\text{quad}}$ is determined by $\theta_{\text{nl}}$ through the H-OpInf algorithm.}
  \label{fig:algorithm}
\end{figure*}
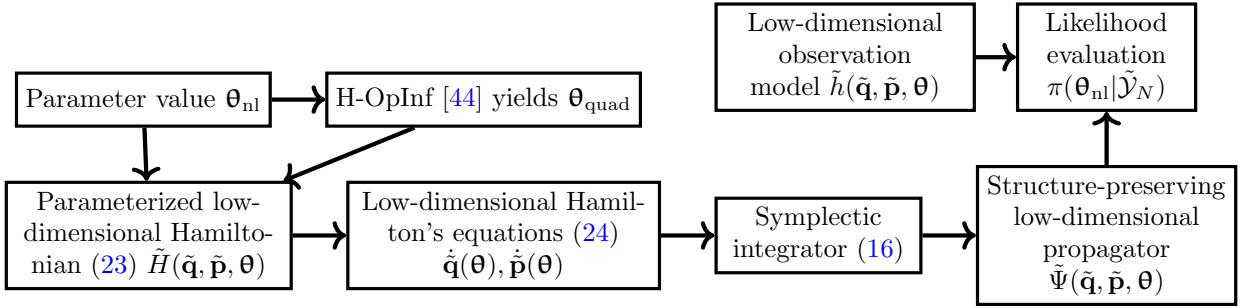

In this section, we discuss the approach of probabilistically learning low-dimensional dynamics of high-dimensional systems, including how to preserve Hamiltonian structure throughout the dimension-reduction process. This approach is divided into three steps. The first step in Section~\ref{sec:lowd_dyn} is defining a structure-preserving propagator in the reduced-dimensional state space, the second step in Section~\ref{sec:lowd_obs} is deriving a low-dimensional observation function that determines a reduced-dimensional observation space, and the final step in Section~\ref{sec:rom_like} is deriving and evaluating a likelihood in this low-dimensional space. This revised process for high-dimensional Hamiltonians is graphically illustrated in Fig.~\ref{fig:algorithm}.

\subsection{Reduced-dimension dynamics}\label{sec:lowd_dyn}
Reducing the dimension of a dynamical system hinges upon a key hypothesis: that the system of interest is a high-dimensional realization of underlying low-dimensional dynamics. Under this assumption, it is \textit{theoretically} possible to estimate dynamics with a considerably lower state dimension than the ambient dimension without any loss of accuracy. Estimating the hypothesized low-dimensional dynamics involves two primary components. The first is identifying a low-dimensional subspace on which a significant portion of the state evolution is contained. The other component is estimating a dynamical model whose state evolves within this low-dimensional space. Here we first describe linear dimension reduction for dynamical systems and then expound on alterations that can be made to preserve symplectic structure.

\subsubsection{Linear dimension reduction}
In this work, we consider linear projections. Let $\phipar\in\reals^{\dimx\times\dimr}$ and $\phiperp\in\reals^{\dimx\times (\dimx-\dimr)}$ be projection matrices that project onto complementary subspaces $\mathcal{S},\mathcal{S}_{\perp}\subseteq\reals^{\dimx}$ with $\dimr\ll\dimx$. Denoting the components of the state $\vx_k$ that lie in $\mathcal{S}$ and $\mathcal{S}_{\perp}$ as $\vxpar$ and $\vxperp$, the state can be written as
\begin{equation}
    \vx_k = \vxpark + \vxperpk 
    = \phipar\phipar^{\top}\vx_k + \phiperp\phiperp^{\top}\vx_k
   = \phipar\vxr_k + \phiperp\vxr_{k,\perp},
\end{equation}
where $\vxr_k\in\reals^{\dimr}$ and $\vxr_{k,\perp}\in\reals^{\dimx-\dimr}$ are the low-dimensional representations of $\vxpar$ and $\vxperp$.

Once the subspace has been identified, the next step is to learn low-dimensional dynamics of the form
\begin{equation}\label{eq:lowd_t}
  \tilde{\T}(\rvxr_k,\rvtheta,\omega) = \tilde{\Psi}(\rvxr_k,\rvtheta) + \tilde{\rvxi}_k(\omega),
\end{equation}
where $\tilde{\Psi}:\reals^{\dimr}\times\reals^{\dimpar}\mapsto\reals^{\dimr}$ is the low-dimensional dynamics propagator and $\tilde{\rvxi}_k(\omega)\in\reals^{\dimr}$ represents the model uncertainty of $\tilde{\Psi}$. Since the form of uncertainty in the model is unknown, we select the additive model for simplicity. We use $\rvxr_k$ as the uncertain state vector and therefore model it as a random vector. The state $\vxr_{k,\perp}$ is omitted from this model as a result of the earlier hypothesis that the dynamics are largely confined to a low-dimensional manifold, in this case $\mathcal{S}$. With these dynamics, predictions of the high-dimensional state can be made using $\rvx_k = \mV\tilde{\Psi}^{k-1}(\rvxr_1,\rvtheta)$, where $\tilde{\Psi}^{k-1}$ denotes $k-1$ compositions of the low-dimensional propagator.

\subsubsection{Hamiltonian case}\label{sec:rom_obs_ham}
Now we seek to extend this methodology to learn low-dimensional Hamiltonian models arising from discretizations of infinite-dimensional Hamiltonian systems. For this work, we focus on infinite-dimensional canonical Hamiltonian systems with Hamiltonian functionals of the form
\begin{equation}
\mathcal{H}[q,p]=\int \left(H_{\text{quad}}(q,p, q_z,p_z,\cdots) + H_{\text{nl}}(q,p) \right) \ \text{d}z,
\label{eq:Hc}
\end{equation}
where $z$ is the spatial variable and $q_z=\frac{\partial q}{\partial z}$ is the partial derivative of $q$ with respect to $z$. The function $H_{\text{quad}}(q,p,q_z,p_z,\cdots)$ contains quadratic terms, and $H_{\text{nl}}(q,p)$ contains spatially-local nonlinear terms. 

To preserve the symplectic structure of the Hamiltonian system, we require that the symplecticity of the Hamiltonian flow is preserved in the reduced-dimensional space. This can be achieved by using the cotangent lift algorithm to find a projection matrix $\phipar$ that is symplectic. We briefly describe the cotangent lift algorithm here, but more details can be found in \ref{sec:cotangent}. Let $\x_1,\dots,\x_{\numsim}$ be a collection of full-field data collected from the system at times $t_1,\dots,t_{\numsim}$. We define the snapshot matrices
\begin{equation}
\mQ=\begin{bmatrix}
       \q_1  \cdots \q_{\numsim}
     \end{bmatrix} \in \reals^{\nz \times \numsim}, \quad 
\Pp=\begin{bmatrix}
      \p_1  \cdots \p_{\numsim} 
     \end{bmatrix} \in \reals^{\nz \times \numsim}.
     \label{eq:snapshot}
\end{equation}
Then we compute a linear symplectic basis 
\begin{equation}\label{eq:cotangent}
\textbf{V}=\begin{bmatrix}
   \mathbf{\mPhi} & \bzero \\
   \bzero & \mathbf{\mPhi} \end{bmatrix}\in \reals^{2\nz \times 2\dimr},      
\end{equation}
where $\mathbf{\mPhi}\in \reals^{\nz \times \dimr}$ is based on the leading $r$ left singular vectors of the extended snapshot matrix $\textbf{X}_e=[\mQ,\Pp]$.

Next, we propose a parameterization of $\tilde{\Psi}$ that preserves the low-dimensional Hamiltonian structure. We define the low-dimensional position and momentum $\vqr,\vpr\in\reals^{\dimr}$ as
\begin{equation}
    \vqr=\mPhi^{\top}\vq,\qquad \vpr=\mPhi^{\top}\vp.
\end{equation}
Based on the Hamiltonian form of Eq.~\eqref{eq:Hc}, we use the parameterization
\begin{equation}\label{eq:Hr_params}
    \Hr(\vqr,\vpr,\rvtheta) = \frac{1}{2}\vqr^{\top}\Dhatq(\rvtheta_{\text{quad}})\vqr + \frac{1}{2}\vpr^{\top}\Dhatp(\rvtheta_{\text{quad}})\vpr + \Hr_{\text{nl}}(\vqr,\vpr,\rvtheta_{\text{nl}}),
\end{equation}
  where $\Hr_{\text{nl}}(\vqr,\vpr,\rvtheta_{\text{nl}})\coloneqq H_{\text{nl}}(\mPhi\vqr,\mPhi\vpr,\rvtheta_{\text{nl}})$ defines the nonlinear terms, and $\Dhatq$ and $\Dhatp$ are symmetric matrices defining the quadratic terms. We have partitioned the parameter vector $\rvtheta=\begin{bmatrix}\rvtheta_{\text{quad}}^{\top}&\rvtheta_{\text{nl}}^{\top}\end{bmatrix}^{\top}$ into components pertaining to the quadratic and nonlinear terms of the Hamiltonian to distinguish between them. Following Hamilton's equations~\eqref{eq:derivative}, the governing equations for this low-dimensional Hamiltonian system are
\begin{equation}
\label{eq:dt_param}
    \dot{\vqr}(\rvtheta) = \frac{\partial\Hr}{\partial\vpr} = \Dhatp(\rvtheta_{\text{quad}})\vpr + \mPhi^{\top}\fq(\mPhi\vqr, \mPhi\vpr,\rvtheta_{\text{nl}}), \qquad
    \dot{\vpr}(\rvtheta) = -\frac{\partial\Hr}{\partial\vqr} = -\Dhatq(\rvtheta_{\text{quad}})\vqr - \mPhi^{\top}\fq(\mPhi\vqr, \mPhi\vpr,\rvtheta_{\text{nl}}).
\end{equation}
Numerically integrating these equations with Tao's integrator as in Section~\ref{sec:integrator} yields a structure-preserving propagator $\tilde{\Psi}$. With the Hamiltonian operator inference method H-OpInf~\cite{sharma2022hamiltonian}, $\rvtheta_{\text{quad}}$ can be determined in a straightforward fashion, leaving only $\rvtheta_{\text{nl}}$ to be estimated. More details on this approach will be given in Section~\ref{sec:rom_like}. 

\subsection{Reduced-dimension observations}\label{sec:lowd_obs}
Now that the state dimension has been reduced, we require an observation function $\tilde{h}$ defined over this reduced-dimensional state. 
Furthermore, if the observations themselves are high-dimensional, the high computational complexity of Gaussian filtering may require reduction of the observations in addition to the states. We describe this approach in the setting of identity observation operators for simplicity of presentation.

To formulate the measurement function in terms of $\rvxr_k$, we first decompose the full-dimensional measurements in terms of $\rvxpar$ and $\rvxperp$. Assuming that the measurement function $h$ in the observation model~\eqref{eq:observations} is the identity, we have
\begin{equation}
    \rvy_k = \rvx_k\odot\rvv_k + \rveta_k
    = \left(\rvxpark + \rvxperpk\right)\odot\rvv_k+\rveta_k
    = \left(\phipar\rvxr_k + \rvxperpk\right)\odot\rvv_k+\rveta_k.
\end{equation}

The next step is to transform $\vy$ into a low-dimensional form. A natural choice for this step is to project $\vy$ onto the subspace spanned by the low-dimensional dynamics to produce the low-dimensional measurements $\vyr = \phipar^{\top}\vy$. This induces the low-dimensional measurement model
\begin{equation}
\label{eq:lowd_m}
    \tilde{\M}(\rvxr_k,\omega) = \phipar^{\top}\rvy_k
      = \phipar^{\top}\Bigl(\left(\phipar\rvxr_k + \rvxperpk\right)\odot\rvv_k \Bigr) + \phipar^{\top}\rveta_k,
\end{equation}
which produces a collection of low-dimensional data $\ynr=\{\phipar^{\top}\vy_k |  k=1,\ldots,N\}$. This gives rise to a modified posterior distribution $\probd(\vtheta\given\ynr)$. In general, this distribution is only an approximation of the original distribution $\probd(\vtheta\given\yn)$, but it can be computed much more efficiently than the original, as we will show in Section~\ref{sec:rom_like}.

In the case of only additive noise, the additive form of the measurement model is preserved
\begin{equation}
  \tilde{\M}(\rvxr_k,\omega) = \rvxr_k + \phipar^{\top}(\rvxperpk + \rveta_k),
\end{equation}
where
\begin{equation}\phipar^{\top}(\rvxperpk + \rveta_k)\sim\N\Big(\phipar^{\top}\Exp{\rvxperpk}, \phipar^{\top}
(\Var{\rvxperpk}+\mcovadd)\phipar\Big)
\end{equation}
can be treated as a non-stationary additive noise term with unknown mean and covariance. This form allows for Gaussian filtering through simple application of the Kalman filter. In the general form of Eq.~\eqref{eq:lowd_m}, however, the noise can no longer be modeled by an additive and multiplicative model, and Eqs.~\eqref{eq:ymean} and~\eqref{eq:yvar} for evaluating the mean and covariance of $\M$ can not be applied to $\tilde{\M}$. Instead, the first two moments of $\tilde{\M}$ are given as
\begin{subequations}\label{eq:exact_moments}
  \begin{align}
    \Exp{\tilde{\M}(\rvxr,\omega)}&= \phipar^{\top}\Big((\phipar\Exp{\rvxr}+\rvxperp)\odot\bar{\vv}\Big), \\
    \Var{\tilde{\M}(\rvxr,\omega)}&=  \phipar^{\top}\Big(\Var{(\phipar\rvxr+\rvxperp)\odot\rvv} + \mcovadd\Big)\phipar. \label{eq:exact_moments_var}
  \end{align}
\end{subequations}
Due to the presence of the unknown $\rvxperp$ term, these moments are not computable. However, $\rvxperp$ can be assumed to be small due to the initial hypothesis that $\rvx$ is mostly contained in $\mathcal{S}$. By neglecting $\rvxperp$, Eq.~\eqref{eq:exact_moments} becomes
\begin{subequations}\label{eq:approx_moments}
\begin{align}
    \Exp{\tilde{\M}(\rvxr,\omega)} &= \Exp{\rvxr}\odot\bar{\vv}, \\
    \Var{\tilde{\M}(\rvxr_k,\omega)} &\approx \Var{\rvxr_k} + \mGamma(\rvtheta),
    \label{eq:approx_moments_var}
  \end{align}
\end{subequations}
where the uncertainty due to $\rvv$ in~\eqref{eq:exact_moments_var} has been absorbed into an estimated stationary additive term $\mGamma(\rvtheta)$ in~\eqref{eq:approx_moments_var}. In this simplified form, both moments can be computed straightforwardly with a standard filtering procedure.

When $\bar{\vv}=\vone$, the approximation~\eqref{eq:approx_moments} is equivalent to assuming the low-dimensional measurement model
\begin{equation}\label{eq:obs_lowd}
  \tilde{\M}(\rvxr_k,\omega) \approx \tilde{h}(\rvxr_k) + \tilde{\rveta}_k,\quad\tilde{\rveta}_k\sim\N(\vzero,\mGamma(\rvtheta)),
\end{equation}
where $\tilde{h}(\rvxr_k)=\rvxr_k$. This model will be shown to yield acceptable accuracy on an example problem in Section~\ref{sec:nlse}.

\subsection{Reduced-dimension likelihood evaluation}\label{sec:rom_like}
Lastly, we require a method for evaluating the likelihood of the low-dimensional Hamiltonian~\eqref{eq:Hr_params}. With the symmetry constraints on $\Dhatq$ and $\Dhatp$, the dimension of $\rvtheta_{\text{quad}}(\omega)$ is $\dimr(\dimr+1)$. This dimension grows polynomially with $\dimr$, and even for $\dimr \ll \dimx$, this scaling can be cumbersome for nonlinear optimization methods. Fortunately, H-OpInf~\cite{sharma2022hamiltonian} provides a method for estimating $\rvtheta_{\text{quad}}$ with linear optimization when $H_{\text{nl}}$ is available. This procedure is described in more detail in~\ref{sec:hopinf}. By treating $\rvtheta_{\text{quad}}$ as a deterministic function of $\rvtheta_{\text{nl}}$ defined through the H-OpInf procedure, the posterior can be simplified as follows
\begin{equation}
    \probd(\vtheta\given\ynr) = \probd(\vtheta_{\text{nl}}\given\ynr)\probd(\vtheta_{\text{quad}}\given\vtheta_{\text{nl}},\ynr)
   = \probd(\vtheta_{\text{nl}}\given\ynr)\delta_{\vtheta_{\text{nl}}}(\vtheta_{\text{quad}}),
\end{equation}
where $\delta_{\vtheta_{\text{nl}}}(\vtheta_{\text{quad}})$ takes the value 1 if $\vtheta_{\text{quad}}$ is computed from $\vtheta_{\text{nl}}$ with H-OpInf and 0 otherwise. Therefore, we only need a method for evaluating the likelihood $\probd(\vtheta_{\text{nl}}\given\ynr)$. The pdf $\probd(\vtheta_{\text{nl}}\given\ynr)$ can be evaluated by applying Algorithm~\ref{alg:marginal} to the HMM from~\eqref{eq:hmm_general} with dynamics $\tilde{\T}$ from~\eqref{eq:lowd_t} and measurements $\tilde{\M}$ from~\eqref{eq:lowd_m}. The full procedure for this evaluation is given in Algorithm~\ref{alg:lowd_learning} and illustrated in Fig.~\ref{fig:algorithm}.

\begin{algorithm}
  \caption{Evaluating the reduced-dimensional likelihood of a high-dimensional Hamiltonian system}
  \label{alg:lowd_learning}
  \begin{algorithmic}[1]
    \Require Full-field data $\yn$
    \Statex\hspace{1.6em} Reduced dimension $\dimr$
    \Statex\hspace{1.6em} Parameter vector $\vtheta_{\text{nl}}$
    \Ensure Evaluation of $\probd(\vtheta_{\text{nl}}\given\ynr)$
    \Statex \hspace{-1em}\textbf{Pre-processing}
    \State Assemble data into snapshot matrices
    {\scriptsize
    \begin{equation*}
        \mQ = \begin{bmatrix}|&|&|&| \\ \vq_1 & \vq_2 & \cdots & \vq_{\numsim} \\ |&|&|&|\end{bmatrix},\quad
        \mP = \begin{bmatrix}|&|&|&| \\ \vp_1 & \vp_2 & \cdots & \vp_{\numsim} \\ |&|&|&|\end{bmatrix}
    \end{equation*}
    }
    \State Compute projection matrix with Eq.~\eqref{eq:cotangent}
    \begin{equation*}
      \phipar\gets\texttt{cotangent\_lift}(\mQ,\mP,\dimr)
    \end{equation*}
    \State Define $\ynr=\{\phipar^{\top}\vy_k | k=1,\ldots,\numy\}$
    \Statex \hspace{-1em}\textbf{Evaluation}
    \State Solve for $\Dhatq(\vtheta_{\text{quad}})$ and $\Dhatp(\vtheta_{\text{quad}})$ using $\vtheta_{\text{nl}}$ with H-OpInf from~\ref{sec:hopinf}
    \State Define low-dimensional propagator $\tilde{\Psi}$ using Tao's integrator~\eqref{eq:exp_symp} and symplectic time derivatives~\eqref{eq:dt_param}
    \State Define low-dimensional observations $\tilde{h}(\rvxr)=\rvxr$
\State Evaluate $\probd(\vtheta_{\text{nl}}\given\ynr)$ using Algorithm~\ref{alg:marginal} with dynamics $\tilde{\T}$~\eqref{eq:lowd_t} and observation model $\tilde{\M}(\vxr)$~\eqref{eq:obs_lowd}
    \end{algorithmic}
\end{algorithm}

\section{Numerical experiments}\label{sec:experiments}
In this section, we study the numerical performance of the proposed Bayesian system ID approach for three Hamiltonian models with increasing levels of complexity. In Section~\ref{sec:nssnn}, we describe an existing method whose model parameterization we implement within the proposed Bayesian approach for comparison. Then in Section~\ref{sec:training}, we give an overview of training considerations that are used throughout each numerical experiment. The final three sections contain the numerical examples. Each experiment and its distinct contribution beyond the authors' previous work~\cite{sharma2022bayesian} is outlined as follows. In Section~\ref{sec:tao's example} we consider a polynomial nonseparable Hamiltonian system from~\cite{tao2016explicit} to demonstrate the advantages of the proposed Bayesian approach compared to a state-of-the-art neural network-based method when the data are sparse and/or noisy. In Section~\ref{sec:double_pendulum} we consider the double pendulum system to show the effectiveness of the proposed approach for non-polynomial, nonseparable Hamiltonian models in the chaotic regime. Finally, in Section~\ref{sec:nlse} we consider high-dimensional data with multiplicative noise from the nonlinear Schr\"odinger equation and employ the novel Algorithm~\ref{alg:lowd_learning} for learning a high-dimensional Hamiltonian model.

\subsection{Parameterization using deep neural networks}\label{sec:nssnn}
The Bayesian approach presented in this paper is flexible enough that it can be paired with any arbitrary parameterization of the dynamics propagator $\Psi$. Here, we choose a recently-developed neural network architecture for learning nonseparable Hamiltonian systems known as the nonseparable symplectic neural network (NSSNN)~\cite{xiong2021nonseparable} as our parameterization. This approach parameterizes the Hamiltonian using a neural network, evaluates Hamilton's equations using auto-differentiation, and integrates the equations with Tao's integrator, see Section~\ref{sec:integrator}. In Sections~\ref{sec:tao's example} and~\ref{sec:double_pendulum}, we parameterize $\Psi$ using the NSSNN and compare the performance of this neural network when it is trained with the negative log posterior as its objective against when it is trained with the objective recommended in its original paper~\cite{xiong2021nonseparable}. The goal of these experiments is to investigate the effect of each objective on the quality of the estimated model when the training set consists of sparse and noisy data.

Here we briefly describe the NSSNN architecture and training procedure from~\cite{xiong2021nonseparable}. The neural network parameterizing the Hamiltonian is composed of six linear layers with sigmoid activation functions following all but the last layer. The training data is a set of input-output pairs. The inputs $(\vq_1^{(j)}, \vp_1^{(j)})$ are a collection of measurements at times $t_1^{(j)}$, and the outputs $(\vq^{(j)}, \vp^{(j)})$ are measurements at times $t^{(j)}=t_1^{(j)}+T$ for $j=1,\ldots \numy$. For each $j$, the input and output are collected from the same trajectory and are separated by a time length of $T$. Since the integrator returns both the physical $\vq$, $\vp$ and fictitious $\bar{\vq}$, $\bar{\vp}$ position and momentum, as described in Section~\ref{sec:integrator}, the loss function contains a term corresponding to each of these variables. An $L_1$ loss is used since it was empirically shown by~\cite{xiong2021nonseparable} to yield better results:
\begin{equation}\label{eq:objective}
    \mathcal{J}(\vtheta) = \frac{1}{\numy}\sum_{j=1}^{\numy}\left\lVert\vq^{(j)}-\hat{\vq}^{(j)}(\vtheta)\right\rVert_1 + \left\lVert\vp^{(j)}-\hat{\vp}^{(j)}(\vtheta)\right\rVert_1 + \left\lVert\vq^{(j)}-\hat{\bar{\vq}}^{(j)}(\vtheta)\right\rVert_1 + \left\lVert\vp^{(j)}-\hat{\bar{\vp}}^{(j)}(\vtheta)\right\rVert_1,
\end{equation}
where the hat $\hat{}$ denotes an estimated value. In the NSSNN training procedure, the weights of each layer are initialized with Xavier initialization, and the Adam optimizer~\cite{kingma2015adam} is used to minimize the loss. The optimizer's hyperparameters (the learning rate and beta values $\beta_1$, $\beta_2$) are problem-dependent and will be reported within each numerical experiment's subsection. Aside from the loss function, we follow the same procedure for training the NSSNN with the negative log posterior. The NSSNN and training procedure are implemented in PyTorch~\cite{paszke2019pytorch} for its auto-differentiation capabilities.

\subsection{Training considerations}\label{sec:training}
In each of these experiments, we consider only single-trajectory training data. To generate the data, we run a numerical solver with a fine timestep $\dt_f$ to achieve high numerical accuracy. The training data are then subsampled from this solution using an interval $\dt_t>\dt_f$. During training, the models use the larger timestep $\dt_t$ for prediction, which helps in improving the computational efficiency at the expense of larger numerical error. Once training is complete, the estimated models are tested by simulating with the fine timestep $\dt_f$. Using a smaller timestep during testing ensures that the learned models have captured the continuous dynamics of the system and not only a discrete mapping at a single timestep~\cite{krishnapriyan2022learning}.

For each of these experiments, we place an improper uniform prior, i.e., constant probability density, on the dynamics' model parameters for the sake of direct comparison with the non-Bayesian method. We do, however, place weakly informative half-normal priors on the process noise variance parameters to enforce positivity and yield better convergence. The output model and the measurement noise variance are assumed to be known and are therefore fixed unless stated otherwise.

\subsection{Tao's example}
\label{sec:tao's example}

The first system that we consider is a one-degree-of-freedom system used in~\cite{tao2016explicit} with the Hamiltonian
\begin{equation}\label{eq:tao_ham}
  H(q,p) = \frac{1}{2}(q^2+1)(p^2+1).
\end{equation}
For this experiment, we seek to assess the performances of the negative log posterior and the $L_1$ objective for training the NSSNN as the training data become fewer and noisier.

\subsubsection{Data generation and training}
We generate a collection of datasets of varying noise level and size for training. To generate the training data, we use Tao's integrator with initial condition $\vx_1=\begin{bmatrix}0&-3\end{bmatrix}^{\top}$ and timesteps $\dt_f=10^{-3}$ and $\dt_t=10^{-2}$. The number of data points $\numy$ in each dataset takes the values $\numy=300,400,\ldots,1000$ corresponding to training intervals of length $\numy\dt_t$. The period for this system is approximately 3.26, so at this $\dt_t$, the dataset timespans range from slightly under a single period to just over three periods. Then, we corrupt the datasets with multiplicative noise $\rvv_k\sim\U[1-a, 1+a]$ for $a=0.00,0.01,\ldots,0.10$. The mean and variance of this noise are $\bar{\vv}=\vone$ and $\mcovmult=\frac{a^2}{3}\mI_{2\dimq}$. The final collection of datasets includes every $(a,\numy)$ combination for a total 88 datasets.

To train the NSSNN, we use an initial learning rate of 0.05 and beta values $\beta_1=0.9$ and $\beta_2=0.999$. Each model is trained for 400 epochs with the learning rate multiplied by 0.8 every 20 epochs. Following the approach presented in the original NSSNN paper~\cite{xiong2021nonseparable}, a batch size of 512 is used with the $L_1$ objective whenever applicable. For the negative log posterior, the process noise variance is parameterized as $\pcovadd=\rtheta_{\pcovadd}\mI_{2\dimq}$. The scalar optimization parameter $\theta_{\pcovadd}$ is initialized at a value of $10^{-3}$, and a learning rate of $0.5\theta_{\pcovadd}$ and beta values $\beta_1,\beta_2=0.1$ are used for optimization. To encourage small values, the prior $\text{half-}\N(0,10^{-12})$ is placed on the random variable $\rtheta_{\pcovadd}$. The positivity constraint is enforced by setting the value of $\theta_{\pcovadd}$ to be 0.9 times its previous value whenever the optimizer places $\theta_{\pcovadd}$ below zero. When $a=0$, we set $\mcovmult=10^{-16}\times\mI_{2\dimq}$ for positive definiteness.

\subsubsection{Time-domain prediction}
Once training is completed on $\numy$ data points, each model is simulated starting at $\vx_1$ for twice the training timespan until $t_{2\numy}$ using timestep $\dt_f$.  The accuracy of the model is then assessed by computing the mean squared error (MSE) defined as
\begin{equation}\label{eq:mse}
    \frac{1}{2\numy}\sum_{k=1}^{2\numy}(\hat{q}_k-q_k)^2 + (\hat{p}_k-p_k)^2.
\end{equation}

The results of the training can vary due to the randomness in the initialization of the NSSNN parameters, in the mini-batching procedure, and in the measurement noise. Sometimes, the optimizer can even get stuck at poor model estimates that yield very high MSE. To mitigate the effect of these outliers, we train 20 models at each $(a,\numy)$ pair and then compare the two methods using the median and minimum MSEs. Each of these 20 rounds of training has different realizations of initial parameters, mini-batches, and measurement noise. We use these particular statistics to quantify two distinct characteristics of the training objectives. The minimum MSE represents the best accuracy that an objective could potentially achieve either through fortuitous initialization and batching or through a more sophisticated optimization methodology. The median MSE, on the other hand, represents the minimum accuracy a user can expect to achieve roughly 50\% of the time using the standard Adam optimizer with a reasonable amount of hyperparameter tuning. Together, these two statistics give a more complete picture of each objective's performance.

\begin{figure*}[ht]
    \centering
    \begin{subfigure}{\linewidth}
        \centering
        \includegraphics[width=0.32\linewidth]{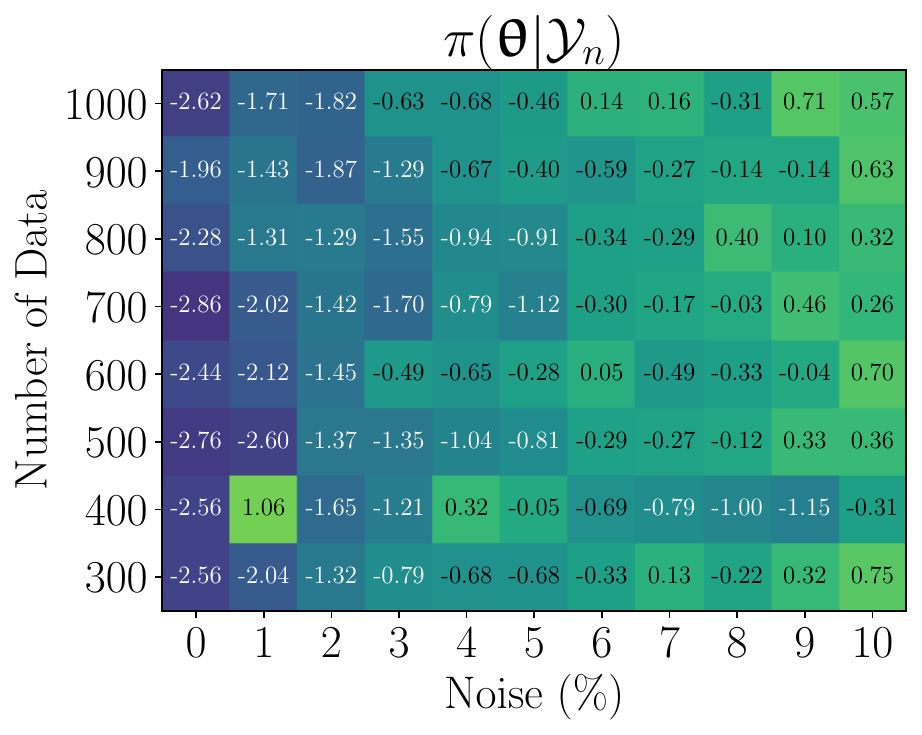}
        \includegraphics[width=0.32\linewidth]{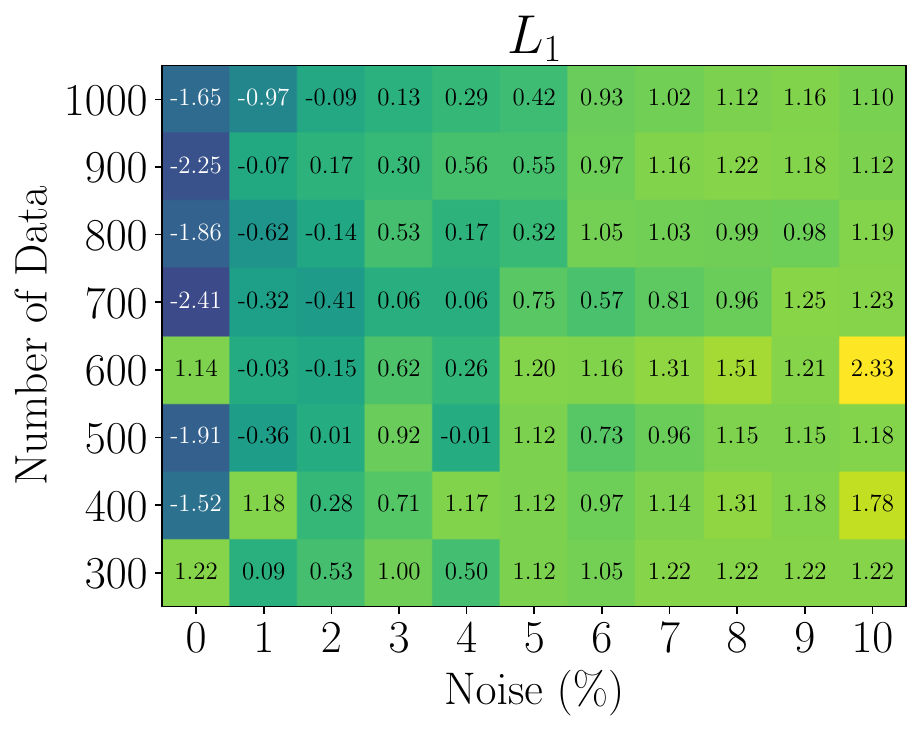}
        \includegraphics[width=0.32\linewidth]{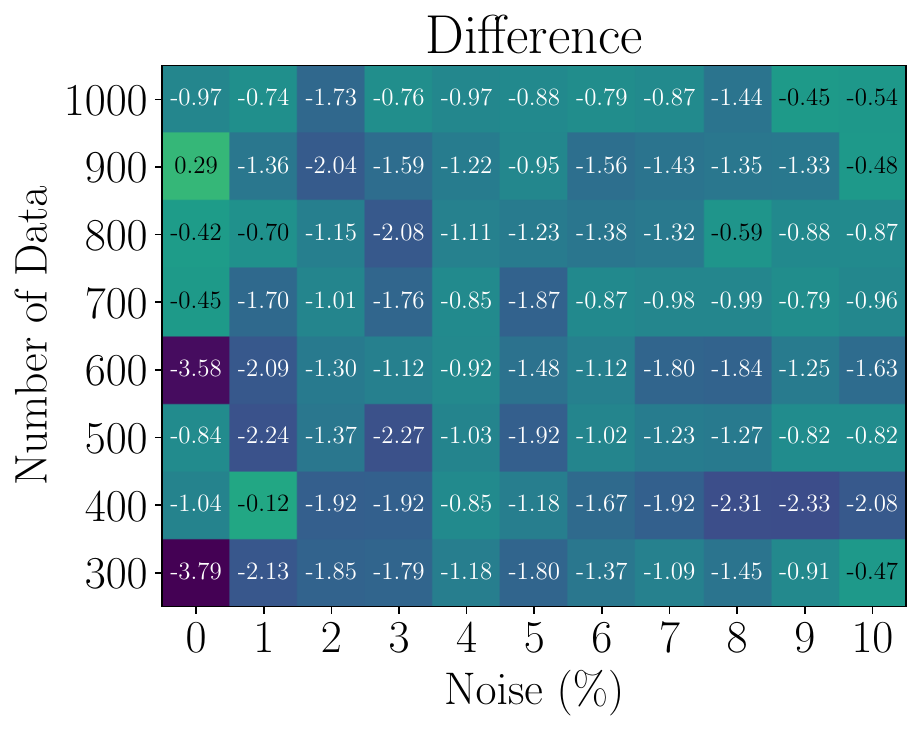}
        \caption{Median}
        \label{fig:median_mse}
    \end{subfigure}
    \begin{subfigure}{\linewidth}
        \centering
        \includegraphics[width=0.32\linewidth]{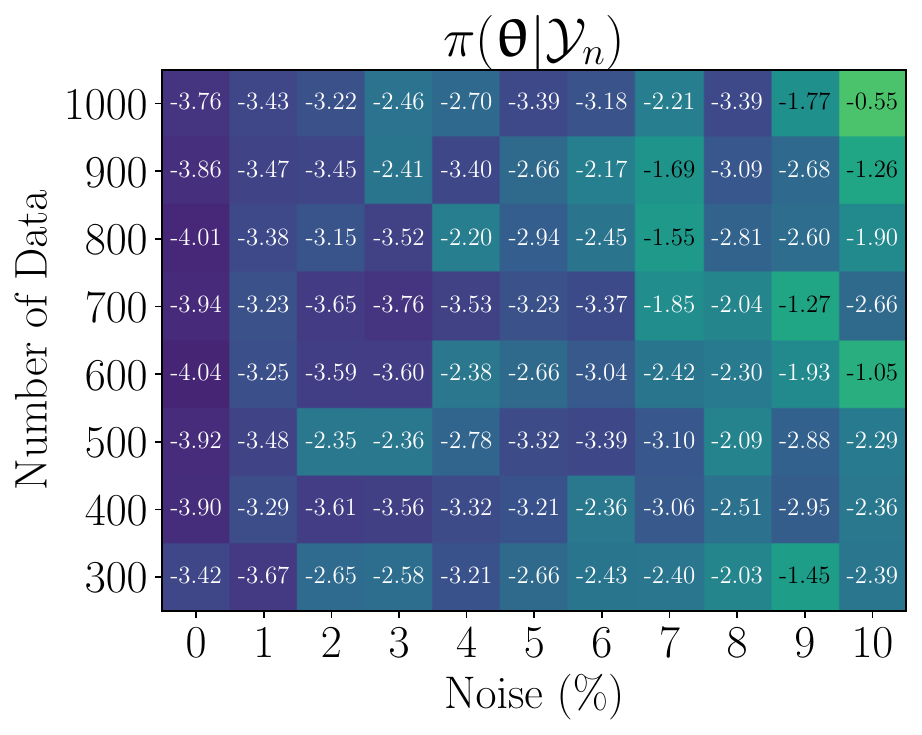}
        \includegraphics[width=0.32\linewidth]{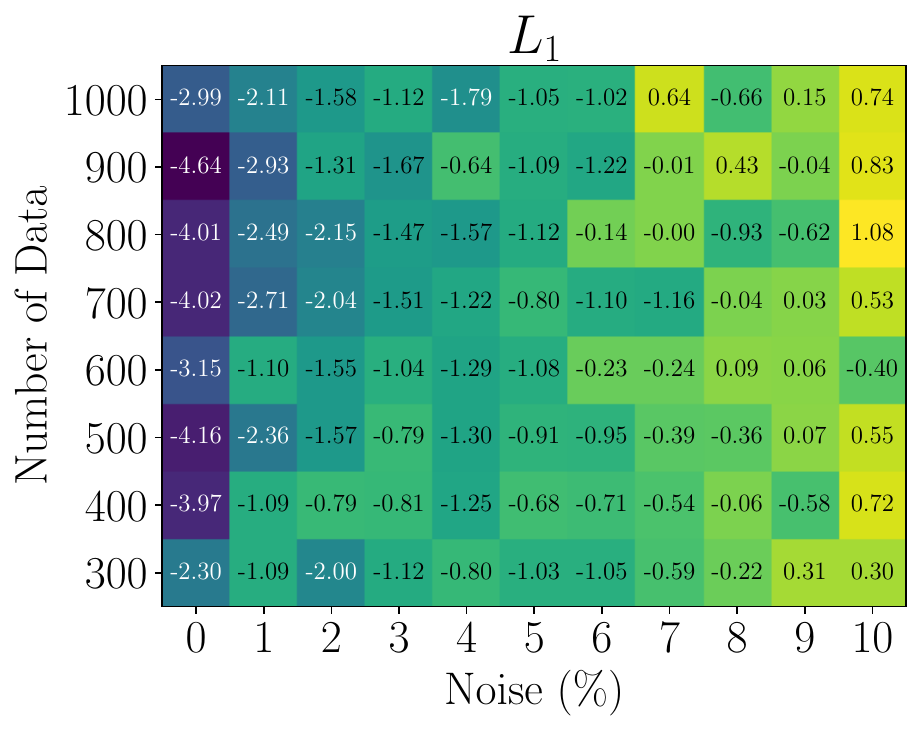}
        \includegraphics[width=0.32\linewidth]{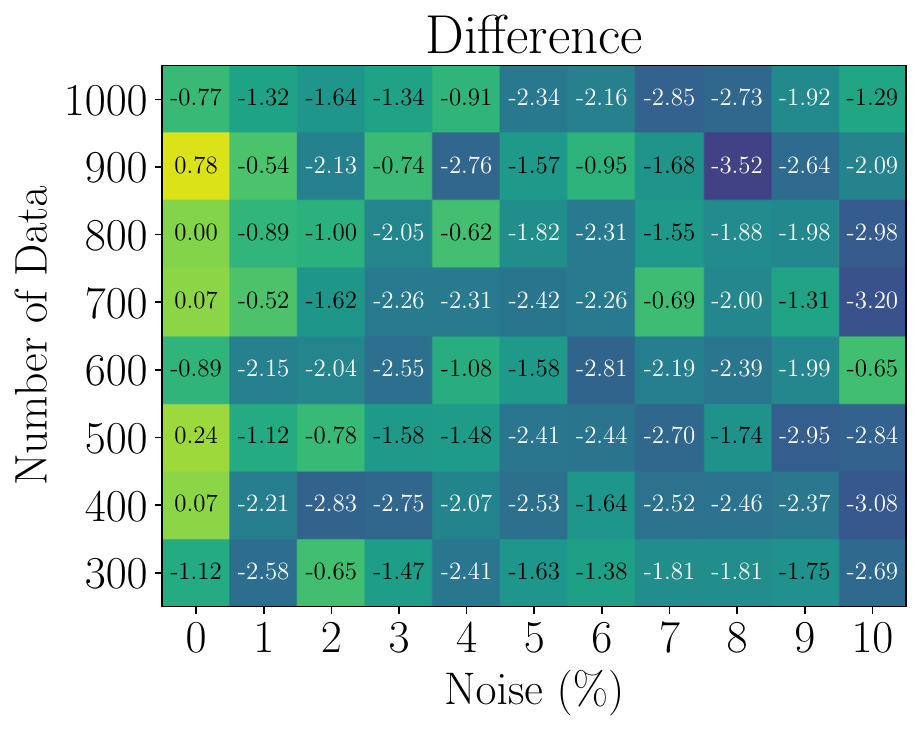}
        \caption{Minimum}
        \label{fig:min_mse}
    \end{subfigure}
    \caption{Tao's example: $\log_{10}$ MSE~\eqref{eq:mse} of models trained using $-\log\probd(\vtheta\given\yn)$ and the $L_1$ norm~\eqref{eq:objective} as objective functions. The label $\probd(\vtheta\given\yn)$ represents the MAP estimate, and `Difference' represents the $\log_{10}$ MSE of the MAP estimate minus the $\log_{10}$ MSE of the $L_1$ estimate. The median MSEs of the MAP estimates are lower than those of the $L_1$ estimates on all datasets, and the minimum MSEs of the MAP estimates are lower than those of the $L_1$ estimates on all datasets with noise.}
    \label{fig:tao_mse}
\end{figure*}

The $\log_{10}$ of the median and minimum MSEs are presented as heatmaps in Figs.~\ref{fig:median_mse} and \ref{fig:min_mse}, respectively. These heatmaps show the effects of varying the number of training data and the level of noise in the data on estimation performance. In this and all future figures, the label $\probd(\vtheta\given\yn)$ denotes the MAP estimate. We observe that using the negative log posterior to train the NSSNN results in lower median and minimum MSE at nearly every $(a,\numy)$ pair. The primary exception is the minimum MSE in the $0\%$ noise column in which the posterior yields lower MSE at only roughly half of the $\numy$ values. As noise is added, however, the MSE of models trained with the $L_1$ loss increases much more steeply compared to the MSE of models trained with the posterior. These results demonstrate that the posterior is much more robust to noise than the $L_1$ loss.  Additionally, the lower median MSE of the posterior in the low noise case shown in Fig.~\ref{fig:min_mse} reflects the fact that the posterior can more reliably find low MSE model estimates, especially when a smaller number of data are available.

\begin{figure}
    \centering
    \begin{subfigure}{0.24\linewidth}
        \centering
        \includegraphics[width=\linewidth]{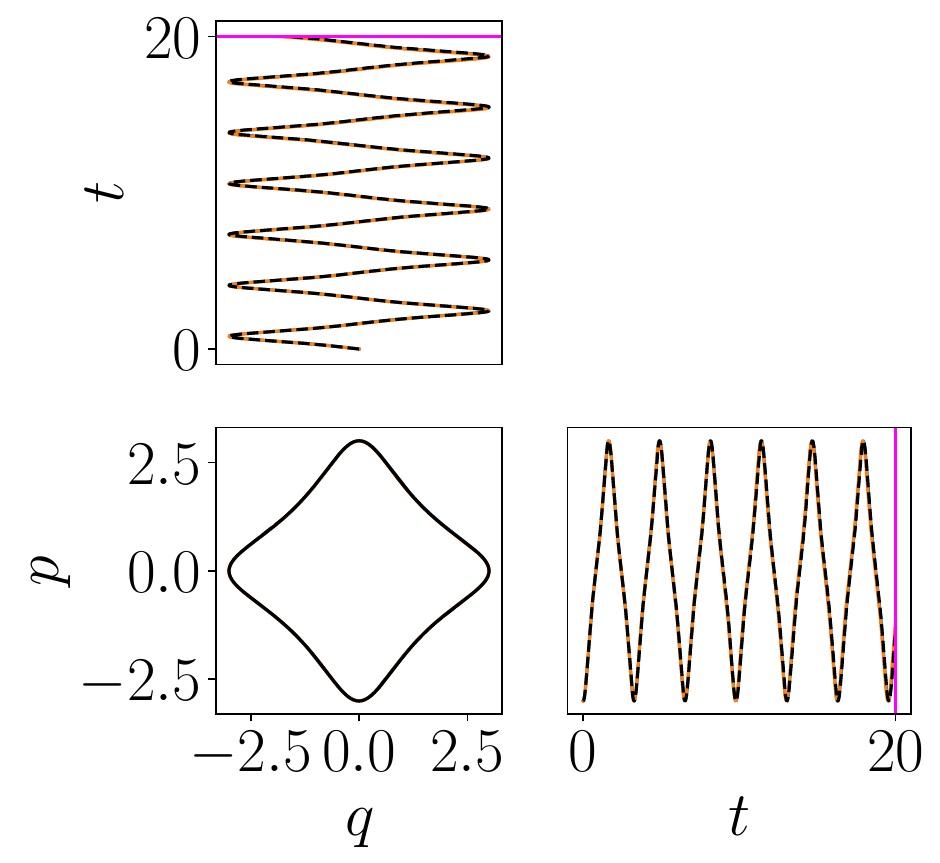}
        \caption{$a=0\%$, $\numy=1000$}
        \label{fig:median_matrix7_0}
    \end{subfigure}
    \begin{subfigure}{0.24\linewidth}
        \centering
        \includegraphics[width=\linewidth]{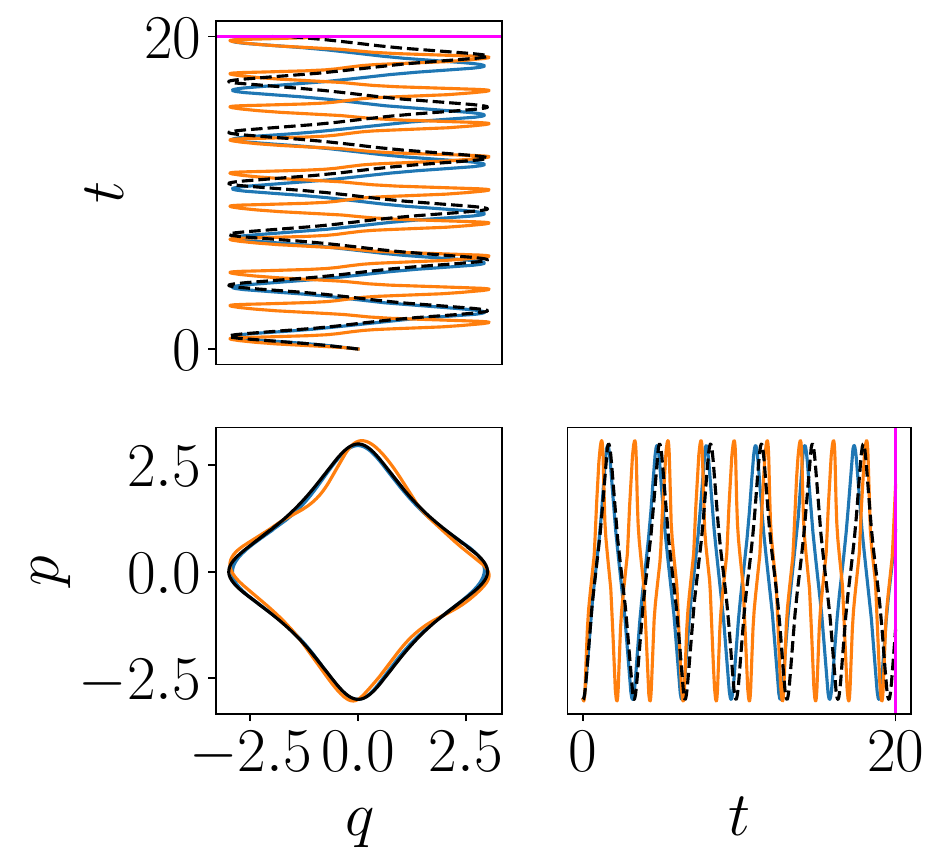}
        \caption{$a=10\%$, $\numy=1000$}
        \label{fig:median_matrix7_10}
    \end{subfigure}
    \begin{subfigure}{0.24\linewidth}
        \centering
        \includegraphics[width=\linewidth]{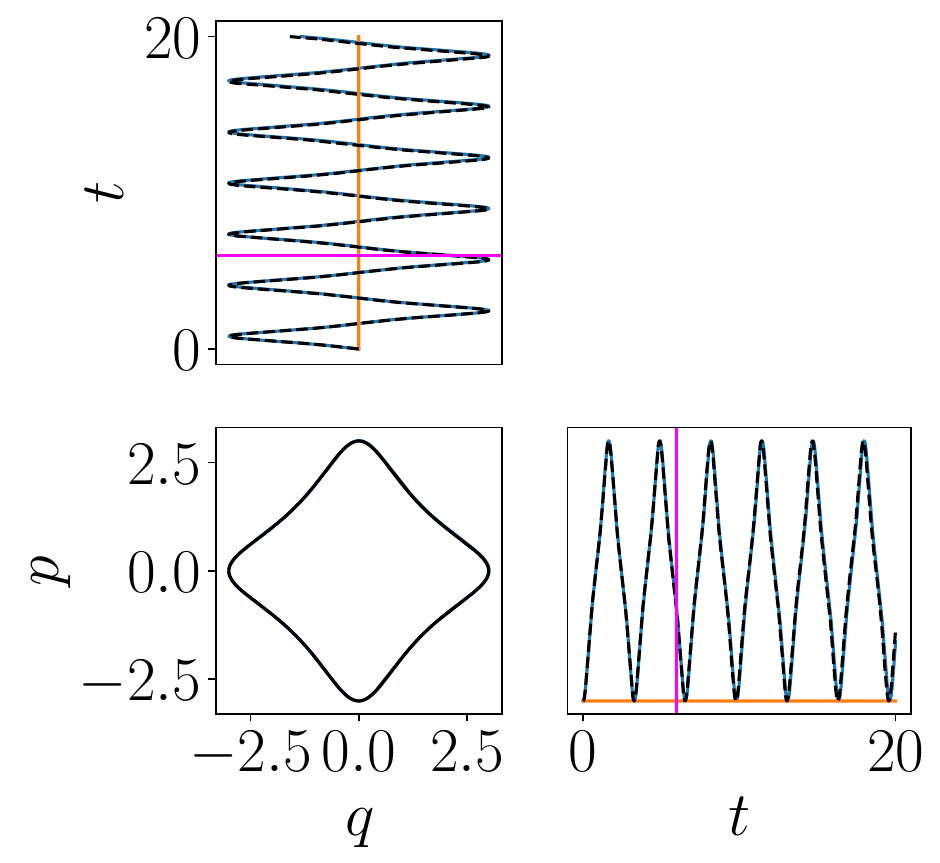}
        \caption{$a=0\%$, $\numy=300$}
        \label{fig:median_matrix0_0}
    \end{subfigure}
    \begin{subfigure}{0.24\linewidth}
        \centering
        \includegraphics[width=\linewidth]{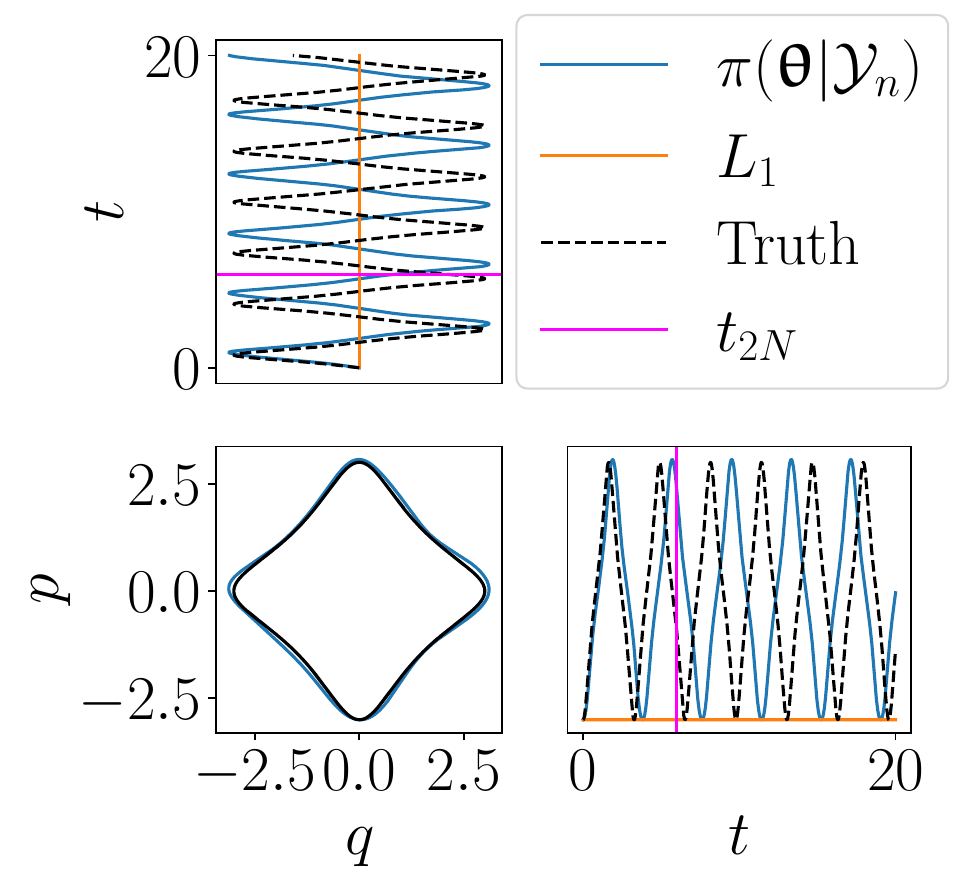}
        \caption{$a=10\%$, $\numy=300$}
        \label{fig:median_matrix0_10}
    \end{subfigure}
    \caption{Tao's example: Estimated trajectories from the median MSE models. The MAP estimate closely matches the truth when the data are noiseless and degrades gracefully as noise increases. The $L_1$ estimates reflect that the optimizer gets caught in local minima when the training dataset is small.}
    \label{fig:median_pred}
\end{figure}

\begin{figure}
    \centering
    \begin{subfigure}{0.24\linewidth}
        \centering
        \includegraphics[width=\linewidth]{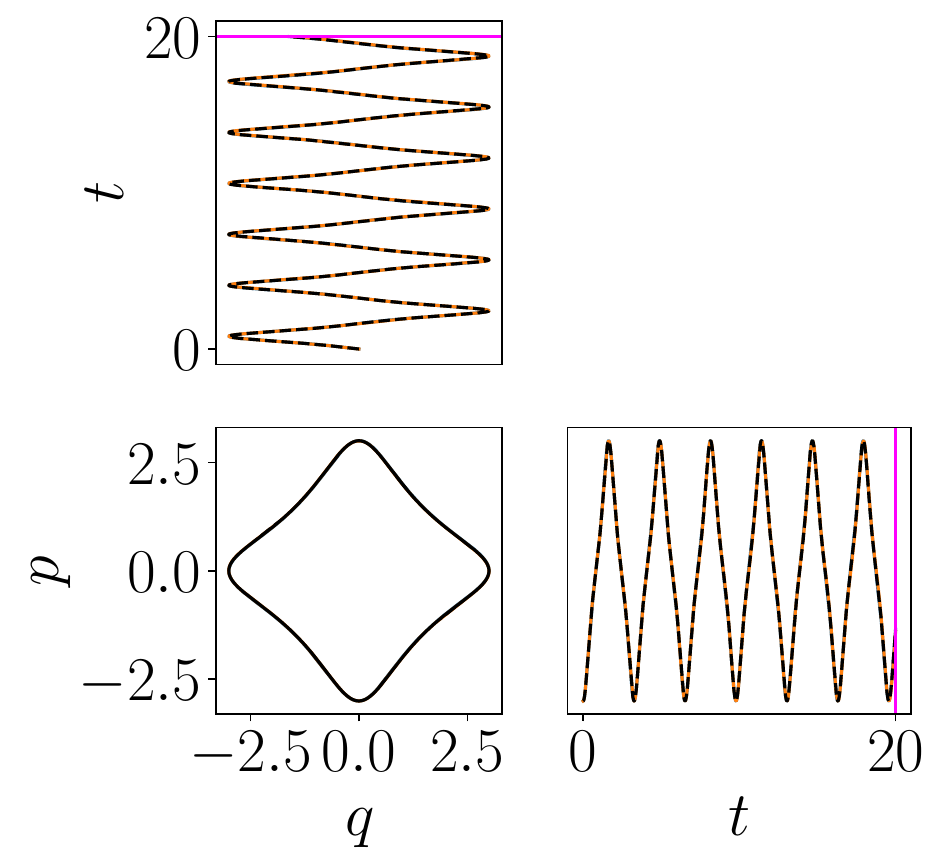}
        \caption{$a=0\%$, $\numy=1000$}
        \label{min_matrix7_0}
    \end{subfigure}
    \begin{subfigure}{0.24\linewidth}
        \centering
        \includegraphics[width=\linewidth]{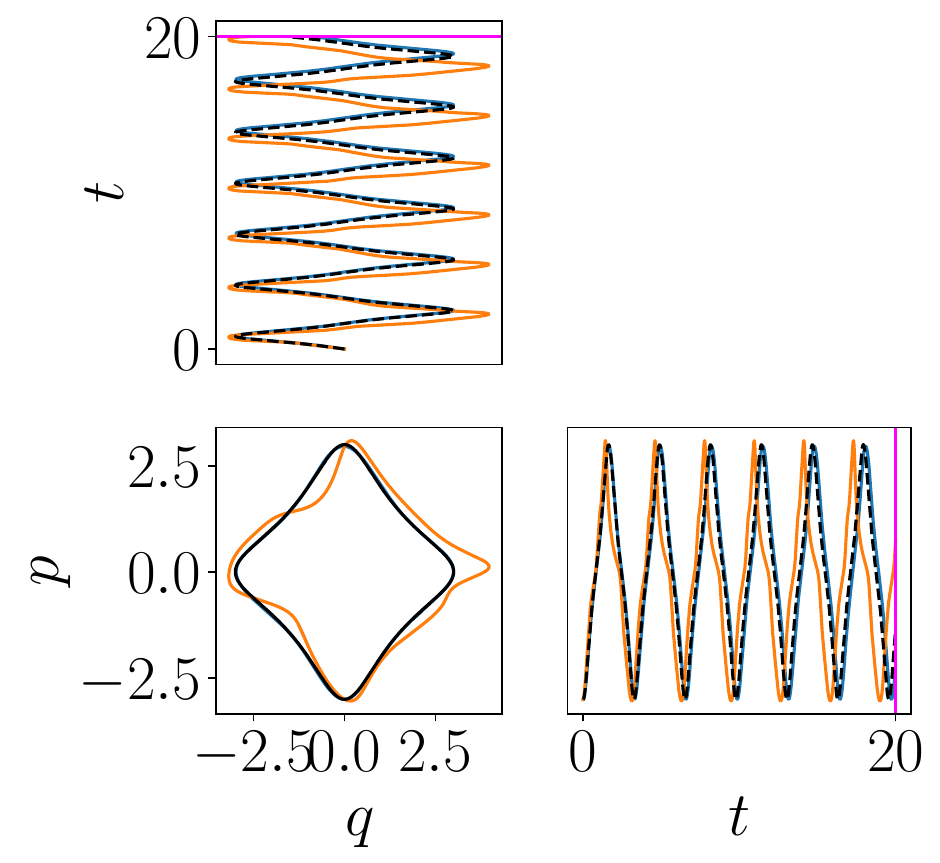}
        \caption{$a=10\%$, $\numy=1000$}
        \label{min_matrix7_10}
    \end{subfigure}
    \begin{subfigure}{0.24\linewidth}
        \centering
        \includegraphics[width=\linewidth]{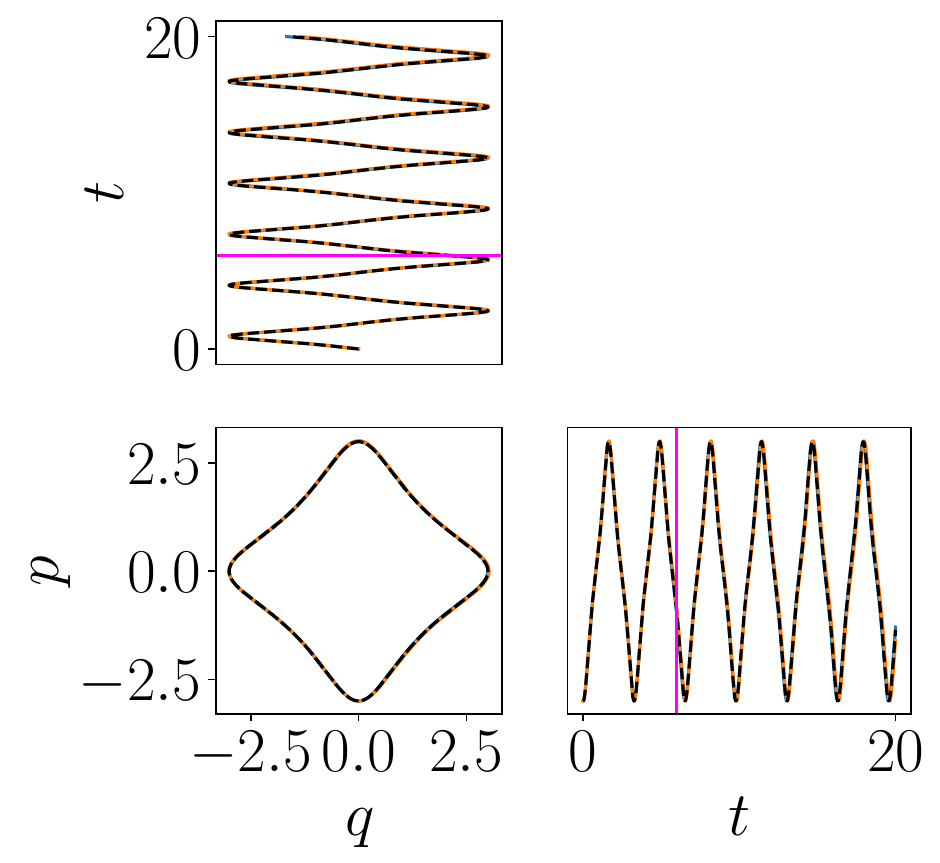}
        \caption{$a=0\%$, $\numy=300$}
        \label{min_matrix0_0}
    \end{subfigure}
    \begin{subfigure}{0.24\linewidth}
        \centering
        \includegraphics[width=\linewidth]{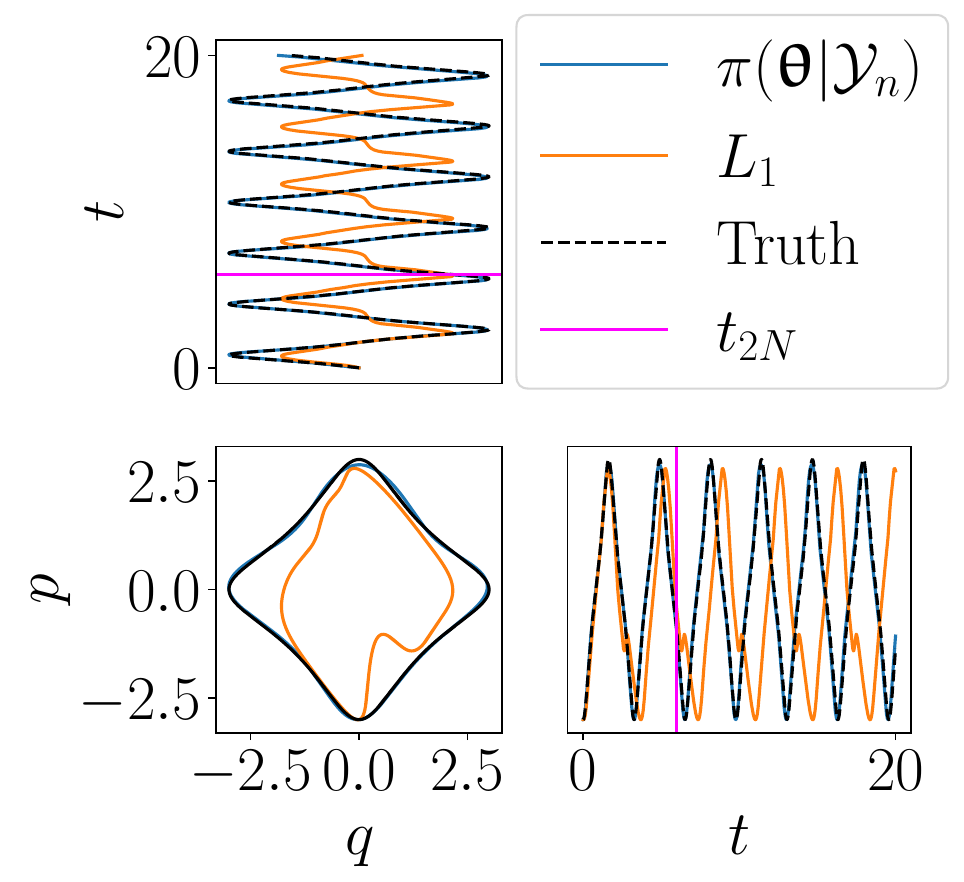}
        \caption{$a=10\%$, $\numy=300$}
        \label{min_matrix0_10}
    \end{subfigure}
    \caption{Tao's example: Estimated trajectories from the minimum MSE models. The MAP estimates match the truth even on noisy datasets, whereas the $L_1$ estimates only match the truth on the noiseless datasets.}
    \label{fig:min_pred}
\end{figure}

To visualize how the model behavior varies with the noise and number of data, we plot the estimated trajectories of models corresponding to the four corners of the heatmaps. The estimated output of the median and minimum MSE models are shown in Figs.~\ref{fig:median_pred} and \ref{fig:min_pred}, respectively. The vertical pink line corresponds to the time $t_{2\numy}$ at which the MSE evaluation ends. Visually, the estimates produced with the negative log posterior can accurately reconstruct the phase manifold in all eight of the figures. The most noticeable inaccuracy produced by the posterior is the oscillation frequency of its median MSE estimates when the noise is high. However, the minimum MSE estimates do not have this issue.

The $L_1$ objective leads to poor results in two main areas: identifying good models with high noise data and reliable optimization with a low number of data. In the high noise case, both the median and minimum MSE estimates produced by the $L_1$ objective construct qualitatively poor estimates of the phase manifold. In the low number of data case, the minimum MSE estimate matches the true output closely when there is no noise. In this same case, however, the median MSE estimate produces a flat line in the time domain. This substantial difference in the median and minimum MSE estimates suggests that although the $L_1$ objective is able to produce good estimates with a small number of data, optimization is challenging and can require multiple attempts.

\subsubsection{Phase-space prediction}\label{sec:tao_phase}
\begin{figure*}[ht]
    \centering
        \includegraphics[width=0.32\linewidth]{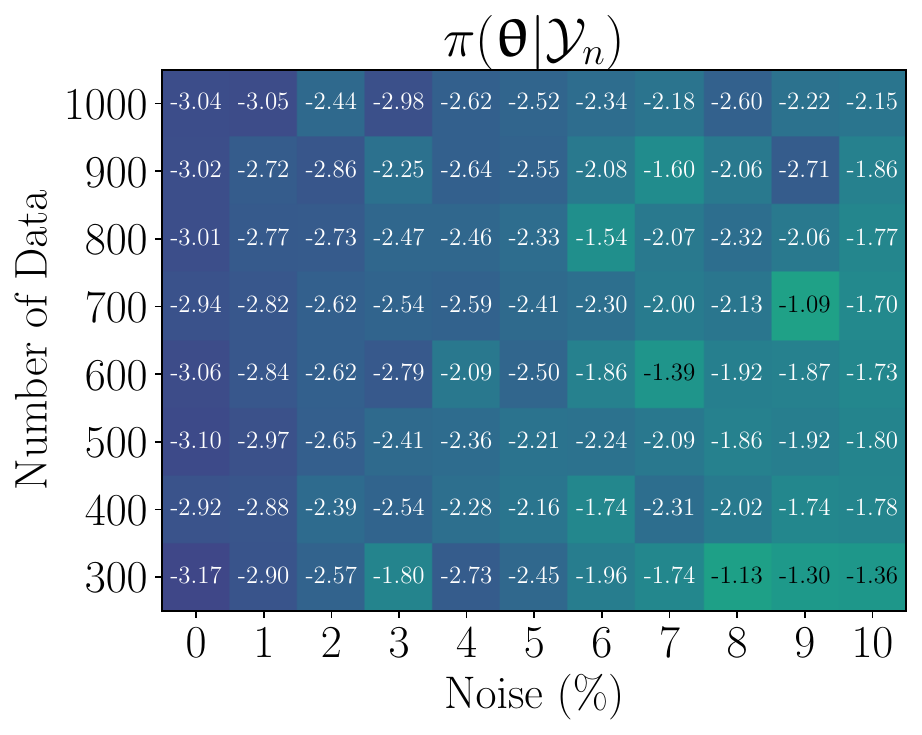}
        \includegraphics[width=0.32\linewidth]{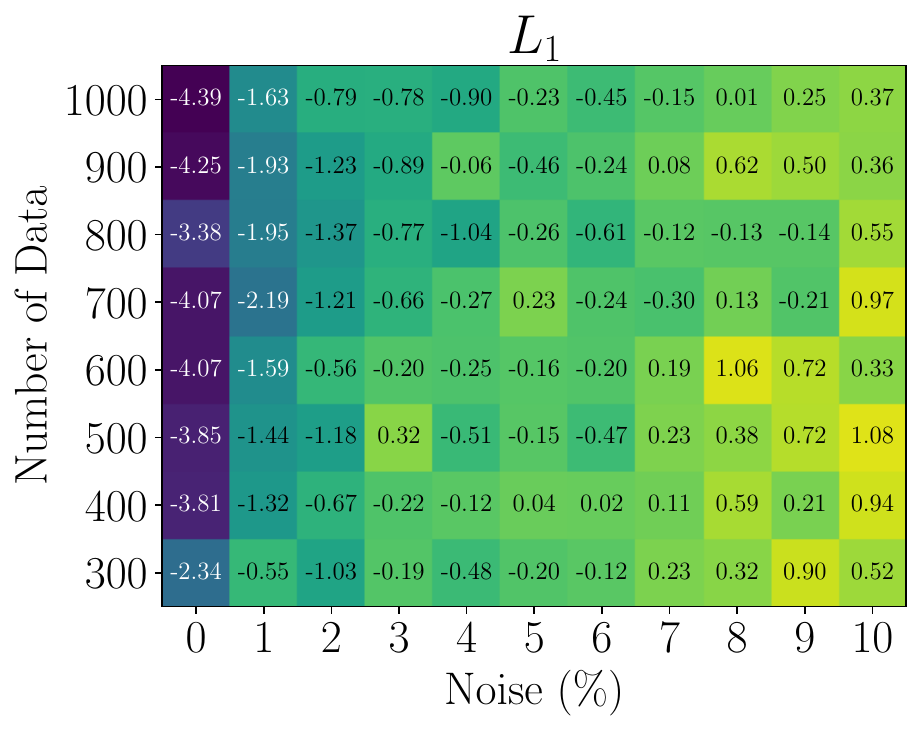}
        \includegraphics[width=0.32\linewidth]{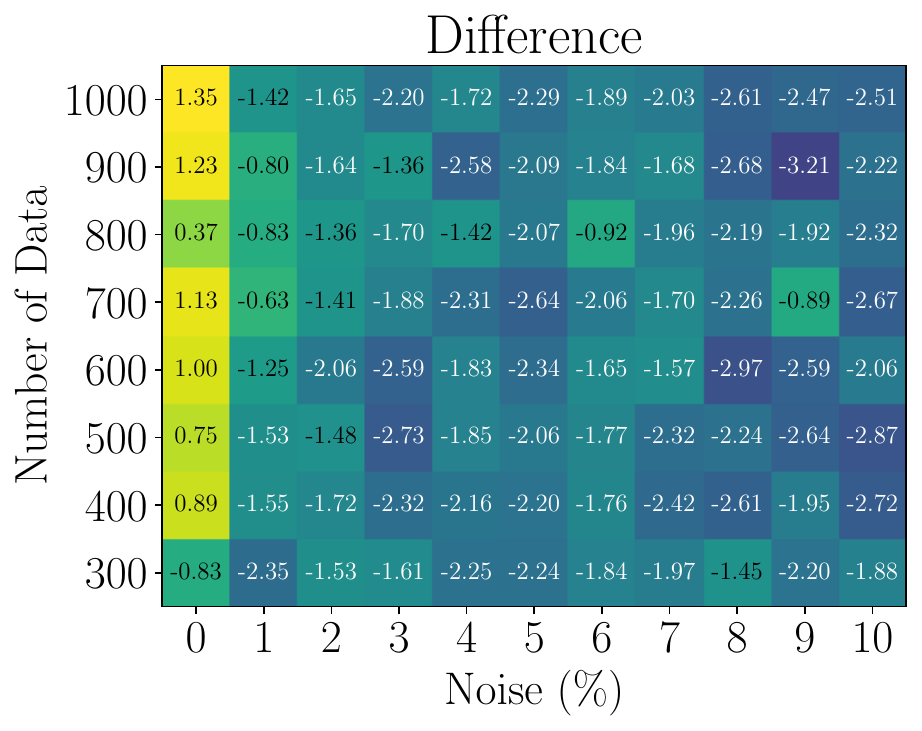}
    \caption{Tao's example: $\log_{10}$ Hamiltonian MSE~\eqref{eq:ham_mse} of the minimum MSE models learned by the $-\log\probd(\vtheta\given\yn)$ and $L_1$ norm objectives. `Difference' represents the $\log_{10}$ Hamiltonian MSE of the MAP estimate minus the $\log_{10}$ Hamiltonian MSE of the $L_1$ estimate. The MAP estimates yield lower Hamiltonian MSE than the $L_1$ estimates on all datasets with noise.}
    \label{fig:ham_mse}
\end{figure*}

In Fig.~\ref{fig:median_matrix7_10}, we noted that the posterior estimate appears to qualitatively match the true output closely in phase space but produces large errors in the time domain. One way to quantitatively assess how closely the estimates match the system behavior in phase space is by looking at the MSE of the Hamiltonian, defined as
\begin{equation}\label{eq:ham_mse}
    \frac{1}{2\numy}\sum_{k=1}^{2\numy}\Big(H(\hat{q}_k,\hat{p}_k) - H(q_1,p_1)\Big)^2,
\end{equation}
where $H$ is the Hamiltonian function~\eqref{eq:tao_ham}. This metric measures the closeness of an estimate's energy level to the true system energy. Since some of the median MSE estimates, such as those in Fig.~\ref{fig:median_matrix0_0}, remain close to the initial point, we only asses the minimum MSE estimates with this metric. The $\log_{10}$ values of the mean squared Hamiltonian deviation for the minimum MSE estimates are shown as heatmaps in Fig.~\ref{fig:ham_mse}. This figure shows a clear trend of the Hamiltonian deviation growing both as noise increases and as the number of data decreases in both objectives' estimates. The estimates produced by the $L_1$ objective outperform those produced by the posterior at almost all $\numy$ values when the data are noiseless --- the notable exception being when the number of data is smallest. 

Although Fig.~\ref{fig:ham_mse} shows that the $L_1$ estimate has a lower Hamiltonian MSE when $a=0$ than the MAP estimate, Fig.~\ref{fig:tao_mse} showed that the MAP estimate yielded generally lower MSE in the time domain for median performance and comparable MSE in the worst-case performance. Moreover the phase-space plots of Fig.~\ref{fig:min_pred} also indicate improved performance of the MAP estimate across noise cases. Looking at the magnitudes of the Hamiltonian MSE values in Fig.~\ref{fig:ham_mse}, the difference in errors between the MAP and $L_1$ estimates at $a=0$ is relatively small, approximately $10^{-3}$. When $a>0$, however, we observe a sharp increase in the Hamiltonian MSE from the $L_1$ estimates. In contrast, the MSE values of the MAP estimates increase gradually as $a$ increases, demonstrating greater robustness to noise.

These slight differences that we observe between the time-domain MSE and Hamiltonian MSE heatmaps arise as a result of the fundamental differences in the two error metrics. While the trajectory-based metric~\eqref{eq:mse} computes the pointwise squared Euclidean distance between two flows parameterized by time, the energy-based metric~\eqref{eq:ham_mse} computes the distance between a flow and a specific energy value. Because the energy-based metric compares the flow to a time-invariant quantity, it cannot by itself properly assess the quality of the dynamics, nor does it gauge or reflect the proper phase-space behavior and level sets. This can lead to problems if, for example, the estimated model possesses a stationary point that coincides with the Hamiltonian level surface. If this stationary point is the initial condition, the flow would yield zero MSE in the energy-based metric while yielding high MSE in the trajectory-based metric. As mentioned earlier, this was the case for some of the median MSE estimates. Although this metric alone is not sufficient to gauge the accuracy of dynamics, it can be valuable when used in conjunction with other metrics for its simplicity and physically-meaningful interpretation. For example, if the energy of a system is known to be constant, the energy-based metric quantifies the physical plausibility of a given model by measuring the average squared change in energy of its flow.

\subsubsection{Conclusions}
We conclude that if the data are noiseless, the $L_1$ objective is the better option for this example because it gives time-domain MSE comparable to the negative log posterior at a lower computational cost. When the number of data is low, however, the negative log posterior is potentially more cost-efficient since it appears to require fewer training restarts than the $L_1$ objective based on the median results.  For data that are noisy, the negative log posterior is the clear choice. Although the cost is higher, the negative log posterior produces significantly more accurate estimates when the data have as little as 1\% noise.

\subsection{Double pendulum}
\label{sec:double_pendulum}
The next system that we consider is the double pendulum, which has the Hamiltonian
\begin{equation}
\small
H(\vq,\vp) = \frac{\mass_2\rod_2^2p_{\phi}^2 + (\mass_1+\mass_2)\rod_1^2p_{\varphi}^2 - 2\mass_2\rod_1\rod_2p_{\phi}p_{\varphi}\cos(q_{\phi}-q_{\varphi})}{2\rod_1\rod_2\mass_2\left(\mass_1 + \mass_2\sin^2(q_{\phi}-q_{\varphi})\right)} -(\mass_1+\mass_2)g\rod_1\cos(q_{\phi}) - \mass_2g\rod_2\cos(q_{\varphi}),
\end{equation}
where $\vq=\begin{bmatrix}q_{\phi} & q_{\varphi}\end{bmatrix}^{\top}$, $\vp=\begin{bmatrix}p_{\phi} & p_{\varphi}\end{bmatrix}^{\top}$, and $\mass_1$ and $\mass_2$ are two masses connected by rigid rods of lengths $\rod_1$ and $\rod_2$. This Hamiltonian is more complex than the previous example, and the behavior displayed by the double pendulum is chaotic in certain regions of the phase space. These two aspects make the system challenging to learn and allow us to test the limits of the NSSNN and the two objective functions.

\subsubsection{Data generation and training}
For this experiment, we set $\mass_1=\mass_2=\rod_1=\rod_2=1$ and $g=9.81$, and we use an initial condition of $\vx_1=\begin{bmatrix}1&0&0&0\end{bmatrix}^{\top}$, placing the dynamics in the chaotic regime. Then, we collect $\numy=2000$ measurements of the full state using timesteps $\dt_f=10^{-3}$ and $\dt_t=10^{-2}$ and corrupt these data with multiplicative noise $\rvv_k\sim\U[0.99, 1.01]$. The training procedure uses 1,000 epochs with an initial learning rate of 0.05 that is multiplied by 0.8 every 50 epochs. To learn the process noise covariance, we use the parameterization $\pcovadd=\diag(\rvtheta_{\pcovadd})$. The learning rate for the optimization parameter $\vtheta_{\pcovadd}$ equals the parameter value and is also multiplied by 0.8 every 50 epochs. Any remaining aspects of the training follow the procedure described in the previous example.

\subsubsection{Time-domain prediction}
\begin{figure*}[ht]
\centering
\includegraphics[width=\linewidth]{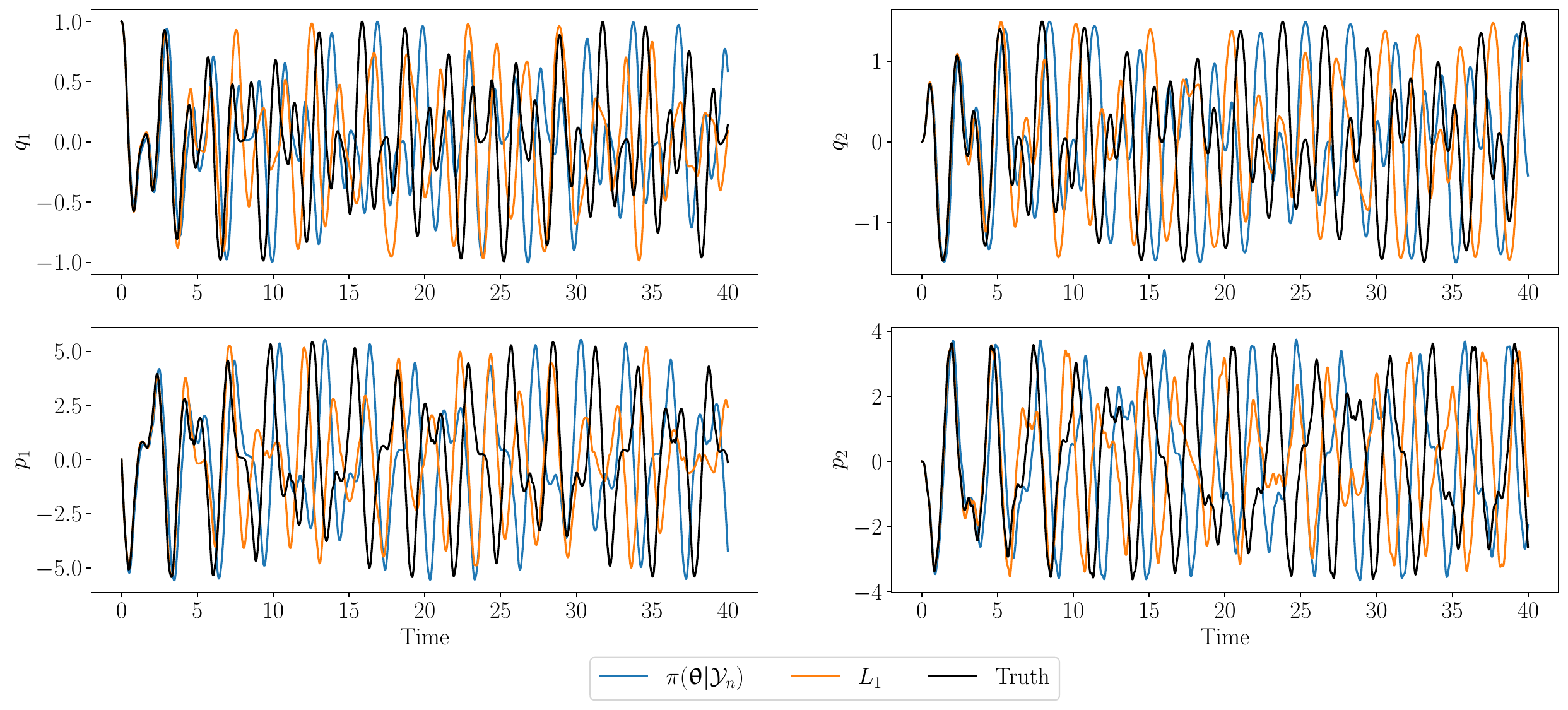}
\caption{Double pendulum: Comparison of the MAP and $L_1$ estimates. Due to the chaotic nature of the system, neither estimate reliably predicts the output for more than five seconds.}
\label{fig:time_point}
\end{figure*}

First, we examine the time-domain estimates of the learned models, plotted in Fig.~\ref{fig:time_point}. Since the double pendulum is a chaotic system, prediction errors grow exponentially fast, making long-term prediction difficult. As a result, the time-domain MSE~\eqref{eq:mse} does not tend to give a reliable quantification of an estimate's accuracy. For chaotic systems, it is often more important to (i) know when the estimate is no longer reliable and (ii) capture the qualitative behavior of the system. 

\begin{figure*}[ht]
\centering
\includegraphics[width=\linewidth]{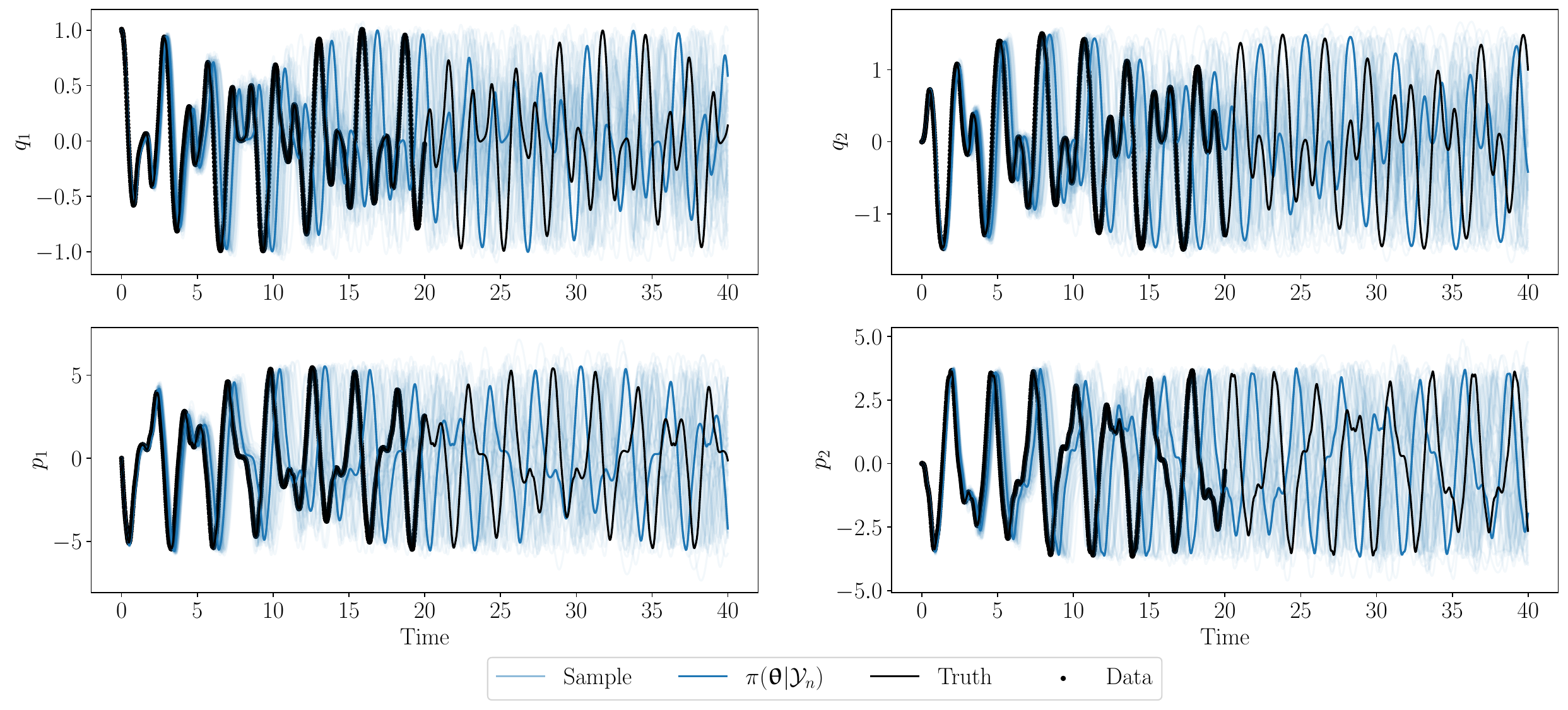}
\caption{Double pendulum: Realizations of the MAP dynamics with the nominal MAP estimate, truth, and data overlaid. The stochastic model approach is able to signal the uncertainty of its estimate through realizations of the estimated process noise.}
\label{fig:time_posterior}
\end{figure*}

To address the first concern, uncertainty quantification can be used. The typical approach to quantifying uncertainty in neural network models is by approximating the posterior distribution over the network parameters~\cite{olivier2021bayesian}, but computational approximation of this distribution is often costly. Fortunately, the proposed Bayesian offers a measure of uncertainty that circumvents this challenging approximation. By including the process noise during evaluation of the model, a stochastic simulation is produced whose realizations can be used to construct a probability distribution of the estimated output. Samples from this stochastic simulation are shown alongside the deterministic simulation, referred to as the ``nominal MAP estimate,'' in Fig.~\ref{fig:time_posterior}. We observe that the spread of samples begins to grow as the MAP estimate begins to deviate from the truth. This suggests that the process noise can give a reliable estimate of how the uncertainty in an estimated model changes over time. Notably, the commonly-used deterministic model approaches have no such resource for assessing the reliability of their estimates.


\subsubsection{Phase-space prediction}
\begin{figure}[ht]
\centering
\begin{subfigure}{0.49\linewidth}
\centering
\includegraphics[width=\linewidth]{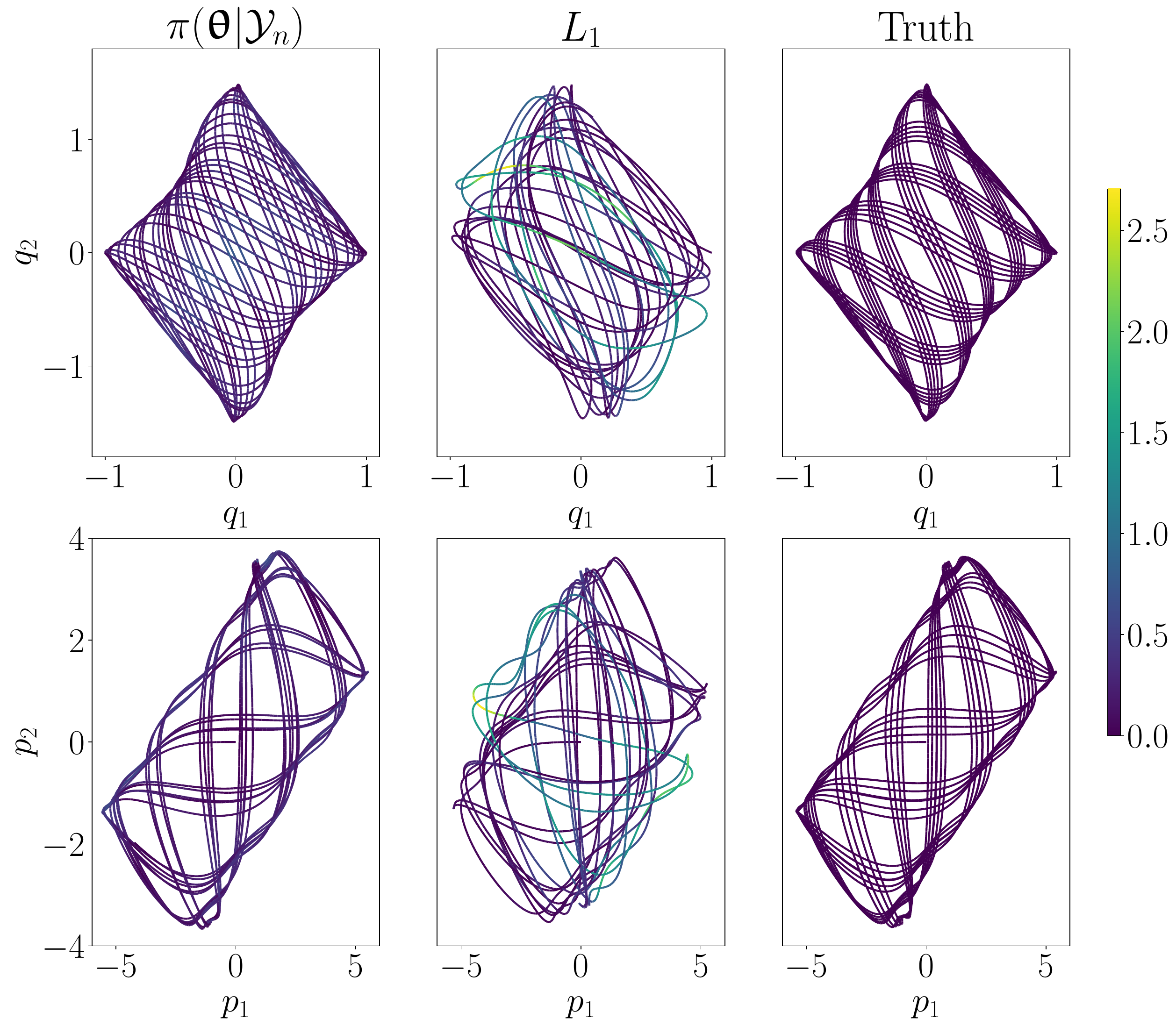}
\caption{MAP and $L_1$ estimates in phase space, where the color denotes the absolute Hamiltonian error. The MAP estimate more closely resembles the shape of the true manifolds.}
\label{fig:phase_point}
\end{subfigure}
\begin{subfigure}{0.49\linewidth}
\centering
\includegraphics[width=0.9\linewidth]{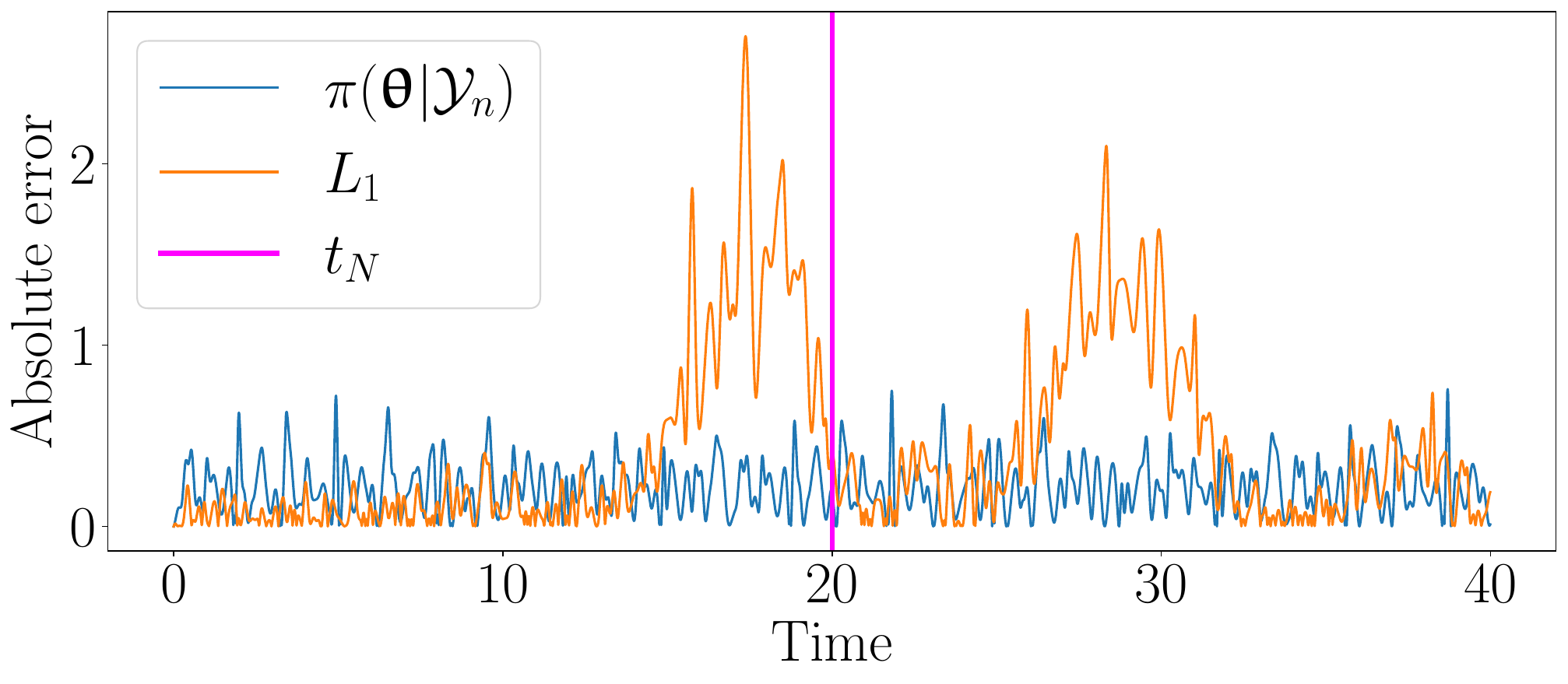}
\caption{Absolute Hamiltonian error over time. The Hamiltonian error of the $L_1$ estimate exhibits greater variance than that of the MAP estimate, indicating overfitting.}
\label{fig:phaseError}
\end{subfigure}
\caption{Double pendulum: Absolute Hamiltonian error in phase (\ref{fig:phase_point}) and time (\ref{fig:phaseError}) domains. The MAP estimate shows lower Hamiltonian error on average.}
\label{fig:abs_ham}
\end{figure}

To address the second concern of assessing qualitative behavior, we compare the accuracy of the two estimates in phase space. Specifically, we quantify this accuracy by evaluating the absolute Hamiltonian error $\left\lvert H(\hat{\vq}_k,\hat{\vp}_k) - H(\vq_1,\vp_1)\right\rvert$. Fig.~\ref{fig:phase_point} shows the estimates in phase space. The color of the line denotes the value of the absolute Hamiltonian error at that point. The absolute Hamiltonian error is also plotted over time for reference in Fig.~\ref{fig:phaseError} with a pink line separating the training and testing time periods. We see that both qualitatively and quantitatively, the MAP estimate is much closer to the true Hamiltonian. Fig.~\ref{fig:phaseError} shows that the error of the $L_1$ estimate is sometimes lower than that of the MAP estimate, but it occasionally spikes to magnitudes that are several times larger than the MAP estimate error. Such behavior is typically indicative of overfitting. Additionally, the spikes become more frequent as time goes on, reflecting the poor potential for long-term forecasting that was also observed in the previous example in Section~\ref{sec:tao_phase}. In contrast, the absolute Hamiltonian error of the MAP estimate is lower on average and does not display sudden spikes. We hypothesize that this preferable behavior can be attributed to the inherent regularization in the marginal likelihood that penalizes large output covariances~\cite{galioto2022likelihood}.

\subsubsection{Conclusions}
On this example, the negative log posterior was better able to capture the underlying Hamiltonian manifold of the chaotic double pendulum in terms of both Hamiltonian error and visual inspection in phase space compared to the $L_1$ objective. Additionally, the Bayesian system ID framework was shown to provide a reliable method of uncertainty quantification of the output forecasts without the expense of quantifying uncertainty in the neural network parameters.

\subsection{Nonlinear Schr\"{o}dinger equation}
\label{sec:nlse}
The nonlinear Schr\"odinger equation (NLSE) is a partial differential equation (PDE) that is used for modeling nonlinear waves in plasma physics~\cite{lan2020nonlinear}, nonlinear optics~\cite{yan2003generalized}, quantum mechanics~\cite{serkin2002exactly}, and oceanography~\cite{akhmediev2009rogue}. This equation exhibits a rich variety of dynamical phenomena due to the combined action of dispersion and nonlinearity on a narrow-banded field envelope. The one-dimensional parametric NLSE considered here is a particularly suitable model for optical experiments realized with single-mode fibers, see~\cite{copie2020physics} for more details.

We consider the parametric nonlinear Schr\"odinger equation with a cubic nonlinearity
\begin{equation}
    \label{eq:npde}
    i\frac{\partial \psi}{\partial t}+ \frac{\partial^2 \psi}{\partial z^2} + \gamma |\psi|^2\psi=0,
    \end{equation}
where $\psi:=\psi(z,t)$ is a complex-valued function and $i=\sqrt{-1}$ is the imaginary unit. This parametric nonlinear PDE can be recast as a canonical Hamiltonian PDE by writing the complex-valued wave function $\psi(z,t)=p(z,t) + iq(z,t)$ in terms of its real and imaginary parts as
\begin{equation*}
     \frac{\partial p}{\partial t}=- \frac{\partial^2 q}{\partial z^2} - \gamma \left( q^2 + p^2 \right)p,\qquad
      \frac{\partial q}{\partial t}=\frac{\partial^2 p}{\partial z^2} + \gamma \left( q^2 + p^2 \right)q,
\end{equation*}
with the space-time continuous Hamiltonian
\begin{equation}\label{eq:nlse_ham}
  \H(q,p)   = \frac{1}{2}\int\left[p_z^2 + q_z^2 - \frac{\gamma}{2}\left(p^2+q^2\right)^2 \right]\rmd z,
\end{equation}
where the parameter $\gamma$ determines the influence of the non-quadratic terms. In addition to Hamiltonian conservation, this system also conserves mass invariant $\Inv_1$ and momentum invariant $\Inv_2$ defined as
\begin{equation}
\label{eq:nlse_inv}
  \Inv_1(q,p) = \int(p^2+q^2)\rmd z, \qquad \Inv_2(q,p) = \int(p_zq-q_zp)\rmd z.
\end{equation}

\subsubsection{Data generation and training}

For the learning problem, we consider a gray-box setting where we assume knowledge about the form of the Hamiltonian~\eqref{eq:nlse_ham} at the PDE level, but the parameter $\gamma$ is uncertain. To differentiate from the true parameter $\gamma$, we denote the uncertain parameter as $\rtheta_{\gamma}$. We then estimate the MAP value of $\rtheta_{\gamma}$ using Algorithm~\ref{alg:lowd_learning} to yield a FOM based on the estimated parameter. To assess the quality of this model, we study the relative error of the estimated states and the model's ability to conserve the system Hamiltonian, mass, and momentum.

\begin{figure}[ht]
    \centering
    \begin{subfigure}{0.49\linewidth}
        \centering
        \includegraphics[width=0.8\linewidth,clip,trim=50 50 0 90]{
          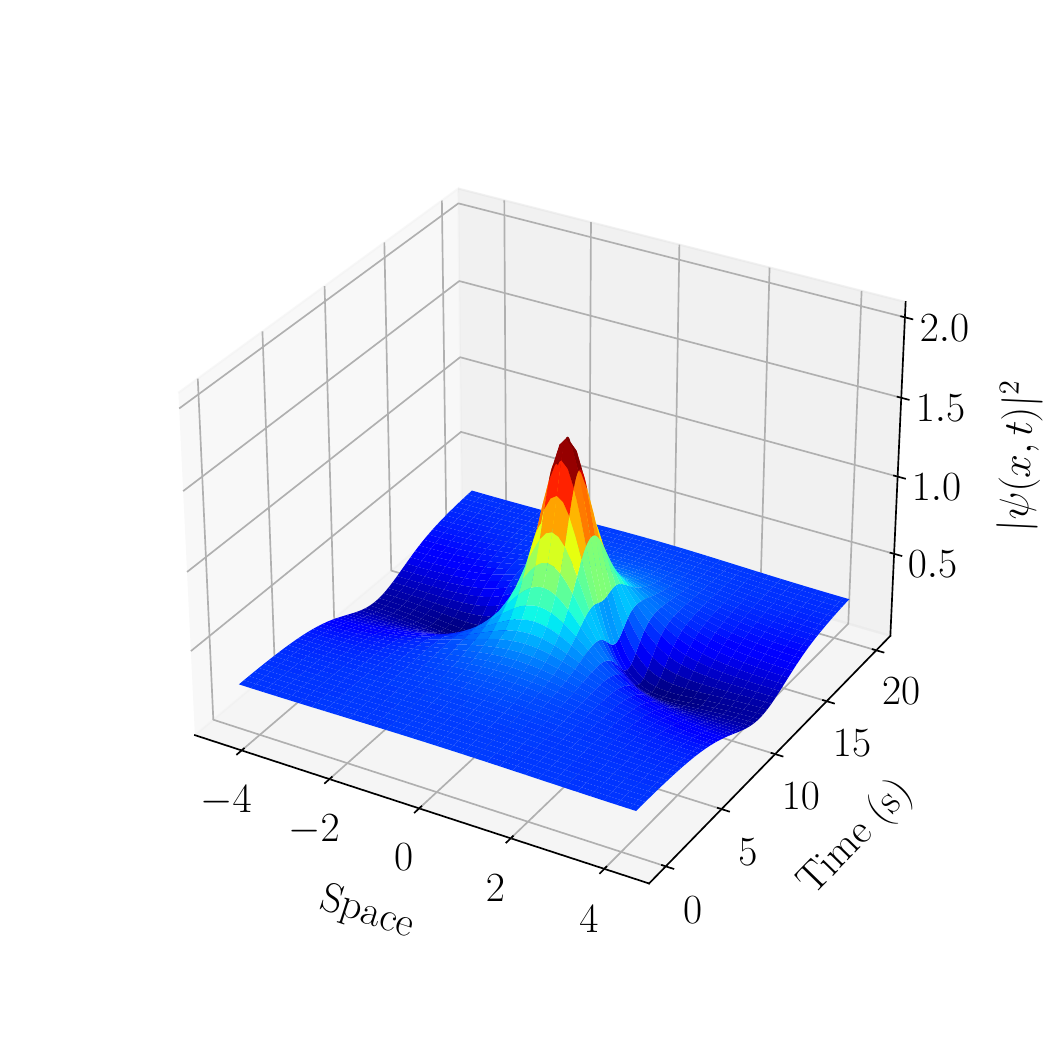}
        \caption{Truth}
        \label{fig:psi_true}
    \end{subfigure}
    \begin{subfigure}{0.49\linewidth}
        \centering
        \includegraphics[width=0.8\linewidth,clip,trim=50 50 0 90]{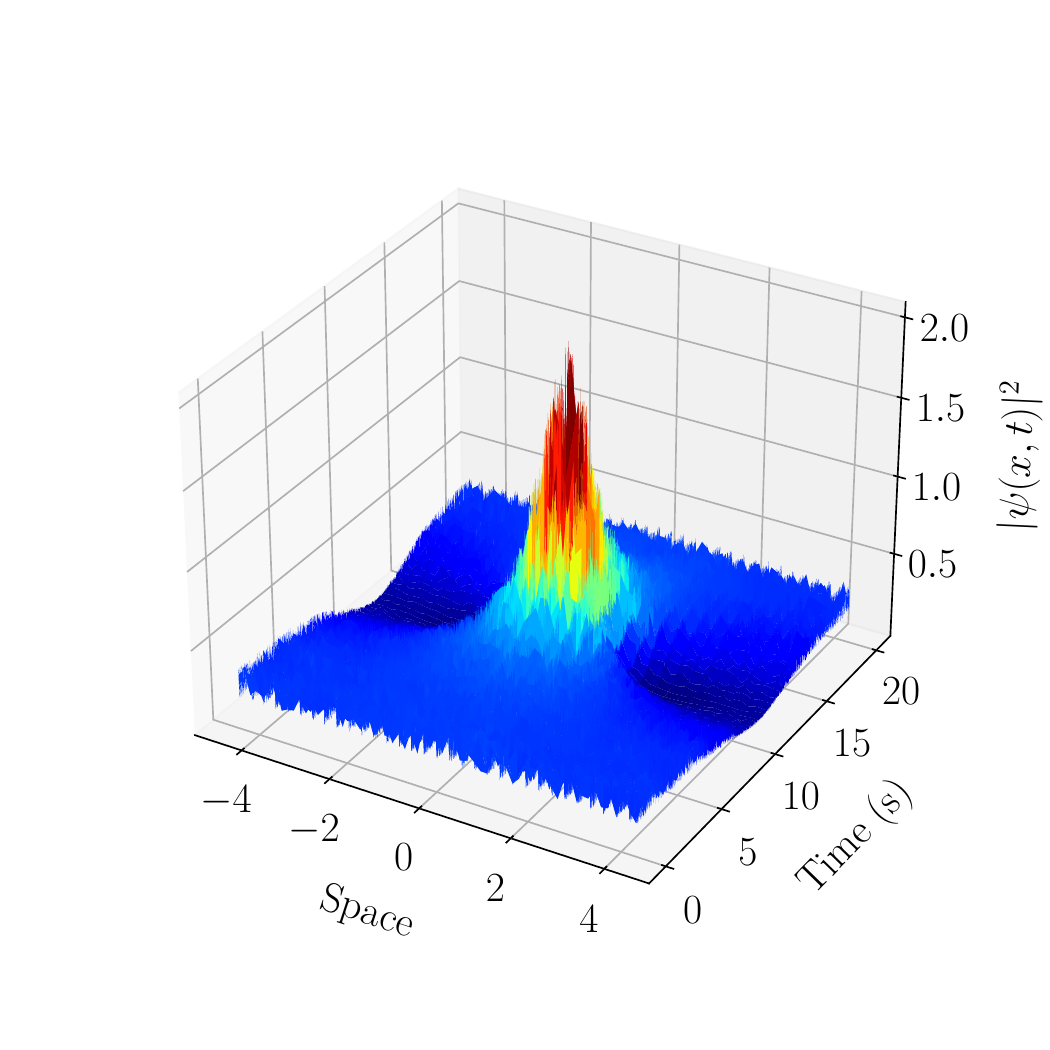}
        \caption{Data}
        \label{fig:psi_data}
    \end{subfigure}
    \caption{NLSE: Data visualization using $\lvert\psi(x,t)\rvert^2$.}
\end{figure}

For this study, we first generate high-dimensional data by numerically solving~\eqref{eq:npde} with true parameter $\gamma=2$. To discretize the PDE in space, we use a spatial domain $z\in[-L/2, L/2]$, where $L=2\pi\sqrt{2}$, with periodic boundary conditions and initial conditions of $p(z,0)=0.5(1+0.01\cos(2\pi z/L))$ and $q(z,0)=0$. The spatial discretization uses $\nz=64$ equally spaced grid points for a total state dimension of $2\nz=128$. We approximate a solution to this discretized PDE using Tao's integrator with a fine timestep of $\Delta t_f=10^{-3}$. Then we collect data over 20s at a timestep of $\Delta t_t=5\times10^{-3}$ for a total $\numy=4000$ data. After generating the data, we add 20\% multiplicative uniform noise $\rvv_k\sim\U[0.8,1.2]$. As a means of visualizing how noisy these data are, we consider the system's wave function $\psi(z,t)$. Among other reasons, the wave function is notable because its square modulus $\lvert\psi(z,t)\rvert^2$ can be interpreted as the probability density of the system. This quantity is plotted using the clean solution in Fig.~\ref{fig:psi_true} and the noisy data in Fig.~\ref{fig:psi_data}.

Next, we project the data onto a low-dimensional subspace so that we can estimate $\rtheta_{\gamma}$ in a reduced setting. For this projection step, we compute a symplectic basis $\textbf{V}$ of the form~\eqref{eq:cotangent} with reduced dimension $\dimr=8$ using the cotangent lift algorithm described in Section~\ref{sec:rom_obs_ham}. As discussed in Section~\ref{sec:lowd_obs}, the multiplicative noise form is not preserved under this dimension-reducing transformation. Therefore, we use the approximate $\tilde{\M}$ given by Eq.~\eqref{eq:obs_lowd} as the observation model for this experiment.

To train the model, we seek to estimate the $\rtheta_{\gamma}$ parameter and the diagonal elements of $\pcovadd$ and $\mcovadd$ for a total of 33 parameters. Since PyTorch does not have a Lyapunov solver needed for H-OpInf, we use the solver from the \texttt{scipy} package, which breaks the computational graph required for auto-differentiation with respect to $\theta_{\gamma}$. As an alternative method of differentiation, we approximate $\partial\probd(\vtheta\given\yn) / \partial\theta_{\gamma}$ using forward finite difference with a step size of $10^{-6}$. Then, we optimize $\theta_{\gamma}$ using gradient descent with a step size of $10^{-4}$. We parameterize the covariance matrices as $\pcovadd=\diag(\rvtheta_{\pcovadd})$ and $\mcovadd=\diag(\rvtheta_{\mcovadd})$. The variance parameters are optimized using the same procedure as before with a learning rate of 0.5 that is multiplied by 0.8 every 10 epochs. The priors over $\rvtheta_{\pcovadd}$ and $\rvtheta_{\mcovadd}$ are $\text{half-}\N(0,10^{-12})$ and $\text{half-}\N(0,10^{-9})$, respectively. We initialize optimization variables $\theta_{\gamma}$ at 0, $\vtheta_{\pcovadd}$ at $10^{-4}$, and $\vtheta_{\mcovadd}$ at $10^{-3}$. For this experiment, we use 50 epochs.

\begin{figure}[ht]
    \centering
    \begin{subfigure}{0.49\linewidth}
        \centering
        \includegraphics[width=0.8\linewidth]{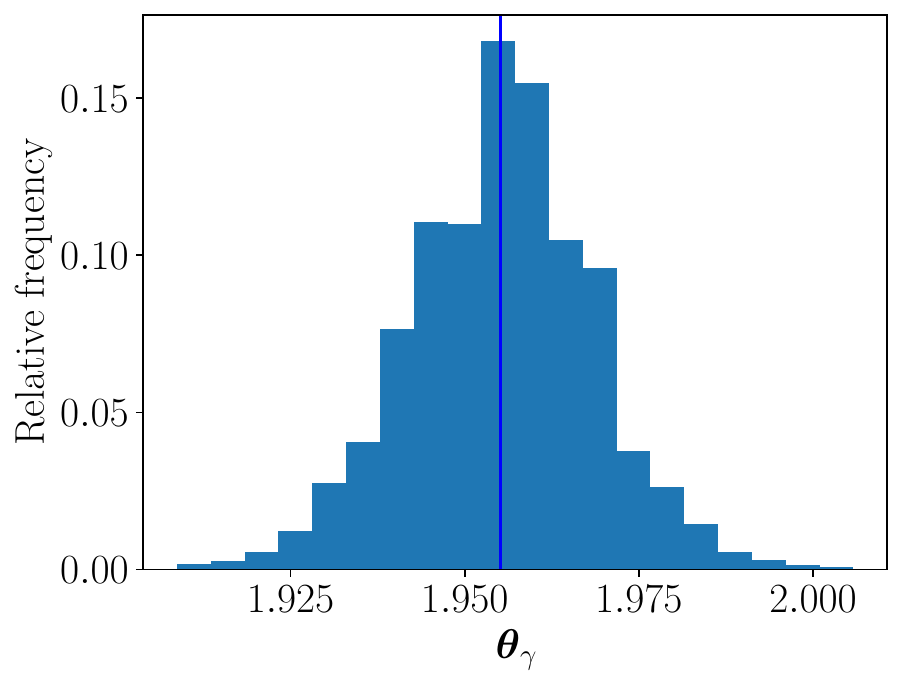}
        \caption{Histogram of samples}
        \label{fig:gamma_hist}
    \end{subfigure}
    \begin{subfigure}{0.49\linewidth}
        \centering
        \includegraphics[width=0.8\linewidth]{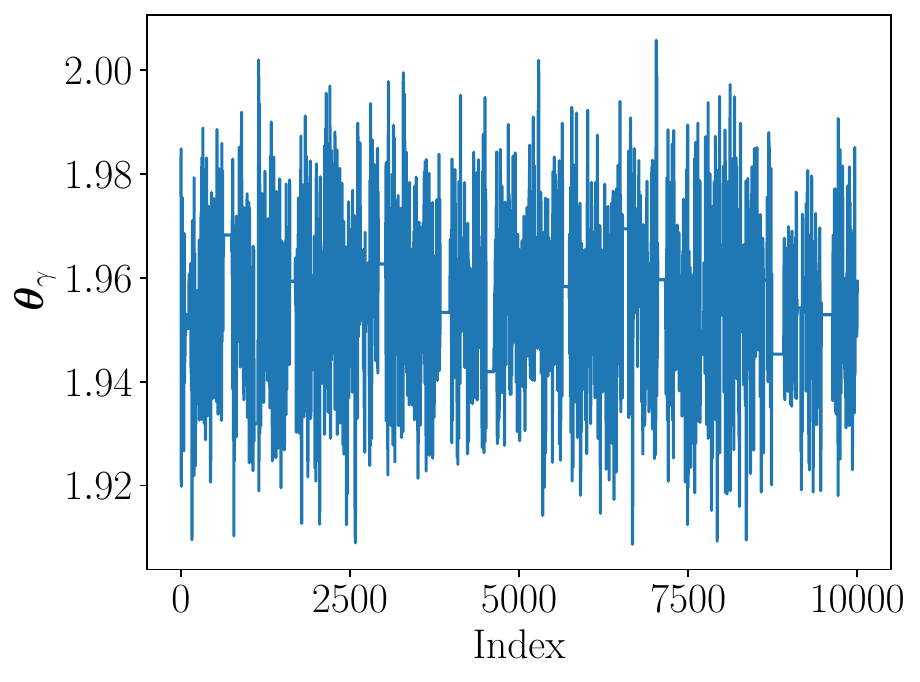}
        \caption{MCMC chain}
        \label{fig:gamma_chain}
    \end{subfigure}
    \caption{NLSE: MCMC samples of the $\theta_{\gamma}$ parameter. The dark blue line in Fig.~\ref{fig:gamma_hist} denotes the parameter posterior mean. The samples are unimodal and the chain appears well-mixed.}
\end{figure}

The optimization result is used to initialize MCMC. The sampling algorithm that we use is a delayed rejection adaptive Metropolis~\cite{haario2006dram} within Gibbs procedure, where the parameter groups $\rtheta_{\gamma}$, $\rvtheta_{\pcovadd}$, and $\rvtheta_{\mcovadd}$ are sampled sequentially. To ensure convergence, we draw $2\times10^{4}$ samples, and discard the first $10^{4}$ as burn-in. Fig.~\ref{fig:gamma_hist} shows the marginal posterior distribution of $\rtheta_{\gamma}$ with the mean value of 1.955 indicated by the dark blue line. The average squared error of these samples with respect to the true $\gamma$ value is $2.185\times10^{-3}$. We also plot the Markov chain of $\rtheta_{\gamma}$ in Fig.~\ref{fig:gamma_chain} to show that the chain is well-mixed.

\subsubsection{Relative state error}

\begin{figure}[ht]
    \centering
    \begin{subfigure}{0.49\linewidth}
        \centering
        \includegraphics[width=0.8\linewidth]{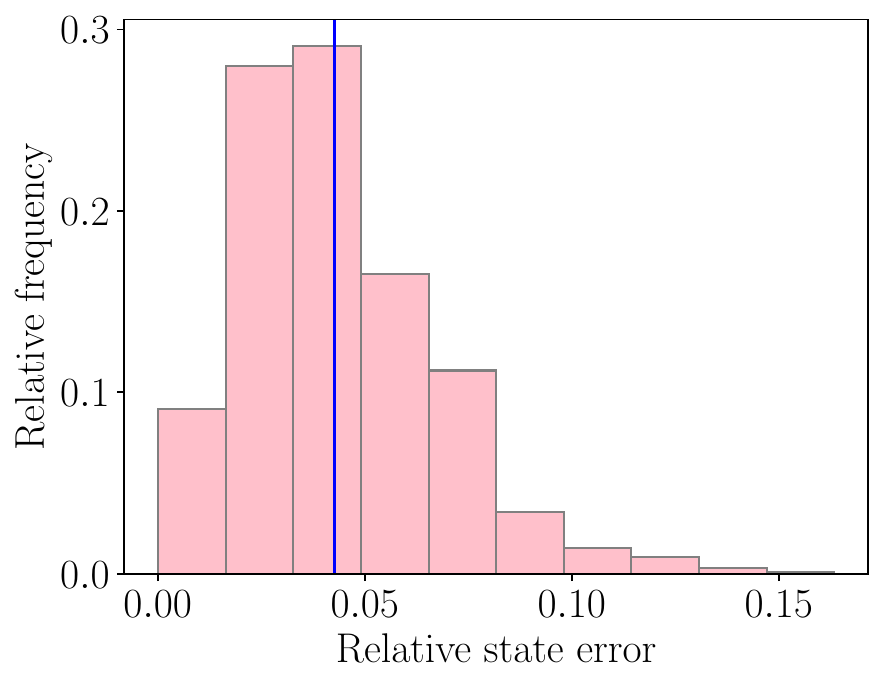}
        \caption{Training period (FOM)}
        \label{fig:state_error_train}
    \end{subfigure}
    \begin{subfigure}{0.49\linewidth}
        \centering
        \includegraphics[width=0.8\linewidth]{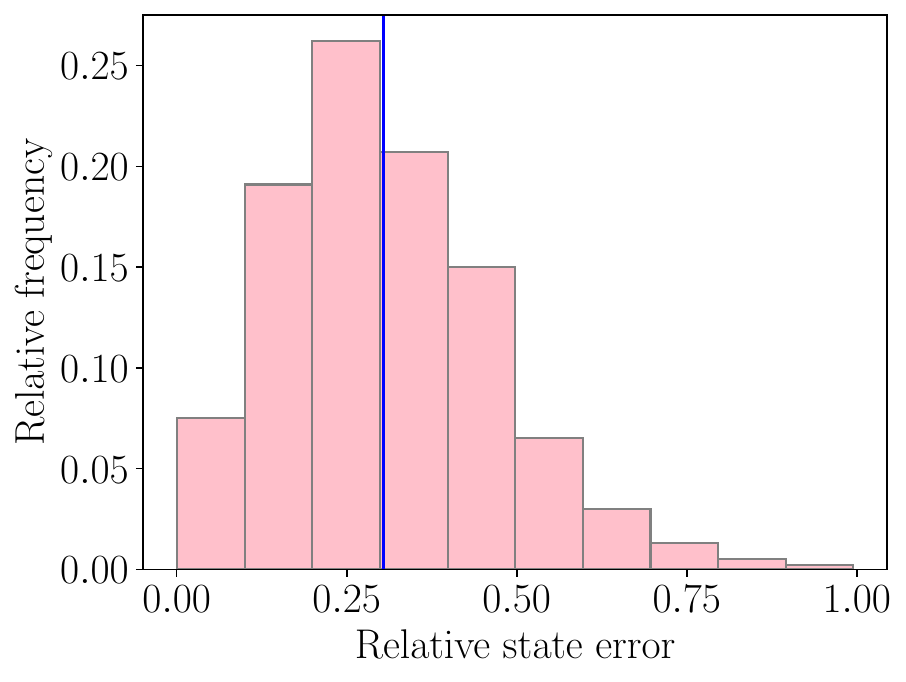}
        \caption{Testing period (FOM)}
        \label{fig:state_error_test}
    \end{subfigure}
    \caption{NLSE: Relative state error~\eqref{eq:rel_state_error} over training and testing periods. The dark blue line denotes the state error of the parameter posterior mean. Most of the samples exhibit less than 10\% state error over the training period. Naturally, the testing period error is higher due to the accumulation of errors over time.}
    \label{fig:state_error}
\end{figure}

We next use these samples to simulate the system over time. To help decorrelate the samples, we only use every 10th sample for a total 1,000 samples. For a given initial condition in the high-dimensional setting, we simulate the learned FOMs until 40s to assess the model performance outside the training period. The performance metric that we consider is the relative state error defined as
\begin{equation}\label{eq:rel_state_error}
  \frac{\lVert \mX_e-\hat{\mX}_e \rVert_F^2}{\lVert \mX_e \rVert_F^2},
\end{equation}
where $\mX_e$ and $\hat{\mX}_e$ are the true and estimated extended snapshot matrices, respectively. The relative state errors over the training and testing periods are shown in Fig.~\ref{fig:state_error}. The dark blue line denotes the relative state error yielded by the $\rtheta_{\gamma}$ sample average. Almost all of the samples of the relative state error over the training period are below 10\%. The error of the testing period is larger due to the fact that errors tend to grow over time in dynamical models. These plots show that we are able to make good predictions over the short-term with errors that grow gradually as the simulation time increases.

\subsubsection{Conservation of invariants}

\begin{figure*}[ht]
    \centering
    \includegraphics[width=\linewidth]{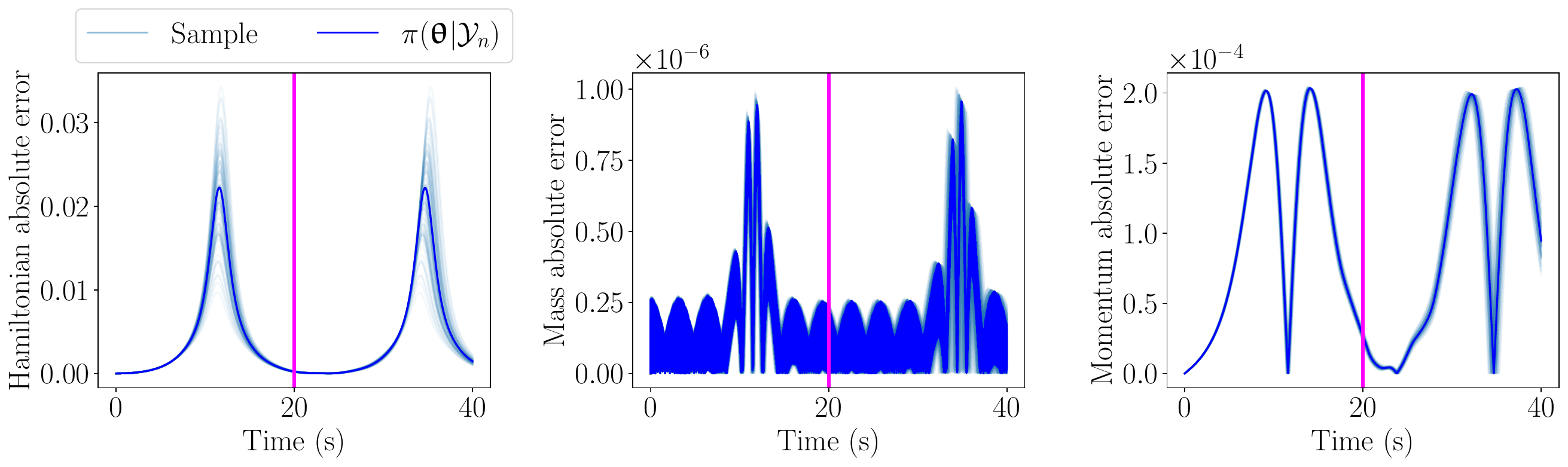}
    \caption{NLSE: Absolute error of the three conserved quantities: Hamiltonian, mass, and momentum. The dark blue line denotes the result yielded by the parameter posterior mean, and the vertical pink line denotes the end of the training period. The sampled FOMs conserve each of the three quantities, demonstrating the structure-preserving capabilities of the proposed algorithm.}
    \label{fig:conservation}
\end{figure*}

We also assess the structure-preserving nature of this model by computing the absolute errors in Hamiltonian, mass, and momentum. The spatially-discretized forms of these invariants from Eqs.~\eqref{eq:nlse_ham} and \eqref{eq:nlse_inv} are defined as
\begin{align}
  H(\vq,\vp) &= \frac{1}{2}\sum_{i=1}^{\nz}\left[p_{z_i}^2 + q_{z_i}^2 - \frac{\gamma}{2}\left(p_i^2+q_i^2\right)^2 \right]\Delta z, \\
  \Invd_1(\vq,\vp) &= \sum_{i=1}^{\nz}\left[p_i^2+q_i^2\right]\Delta z, \\
  \Invd_2(\vq,\vp) &= \sum_{i=1}^{\nz}\left[p_{z_i}q_i - q_{z_i}p_i\right]\Delta z,
\end{align}
where $\Delta z = L/\nz$, $p_{z_i}=(p_{i+1}-p_i) / \Delta z$, and $q_{z_i}$ is defined similarly. The posterior of the absolute errors of the Hamiltonian $\lvert H(\hat{\vq}_k,\hat{\vp}_k)-H(\vq_1,\vp_1)\rvert$, the mass $\lvert \Invd_1(\hat{\vq}_k,\hat{\vp}_k)-\Invd_1(\vq_1,\vp_1)\rvert$, and the momentum $\lvert \Invd_2(\hat{\vq}_k,\hat{\vp}_k)-\Invd_2(\vq_1,\vp_1)\rvert$ are plotted in Fig.~\ref{fig:conservation}. The FOM based on the estimated parameter value exhibits bounded error behavior $100\%$ outside the training data regime for all three conserved quantities, demonstrating that this method is capable of preserving underlying physics.

\subsubsection{Varying the reduced dimension}
\begin{figure}[ht]
        \centering
        \includegraphics[width=0.4\linewidth]{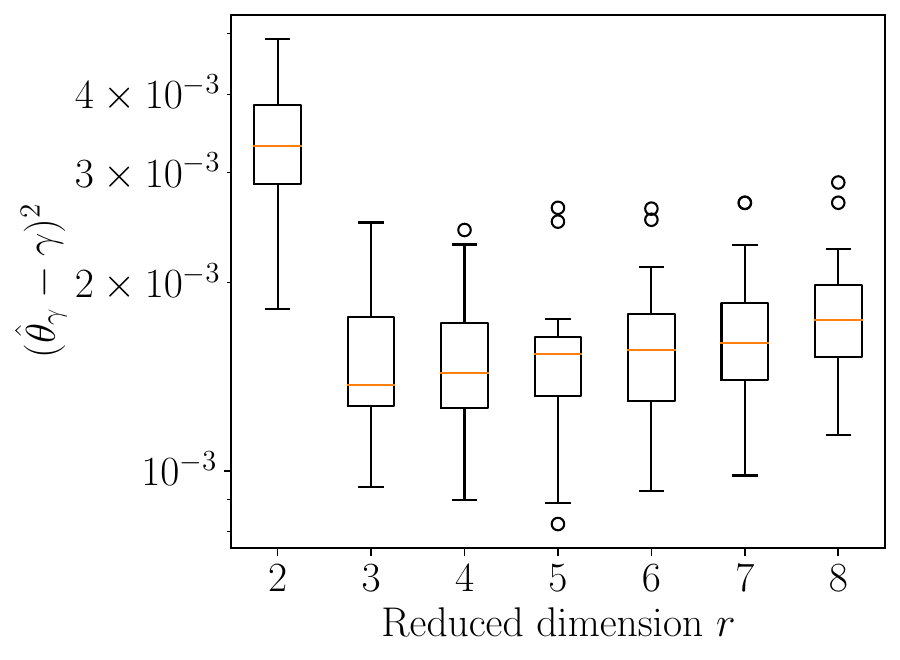}
        \caption{NLSE: Squared error of the parameter MAP $\hat{\theta}_{\gamma}$ at various $\dimr$ over 20 realizations of measurement noise. The circles represent outliers. The errors level off at $\dimr=3$, showing that only three dimensions are needed to achieve the best parameter estimation accuracy with the proposed algorithm on this example.}
        \label{fig:param_error_vs_r}
\end{figure}

\begin{figure}[ht]
        \centering
        \includegraphics[width=0.4\linewidth]{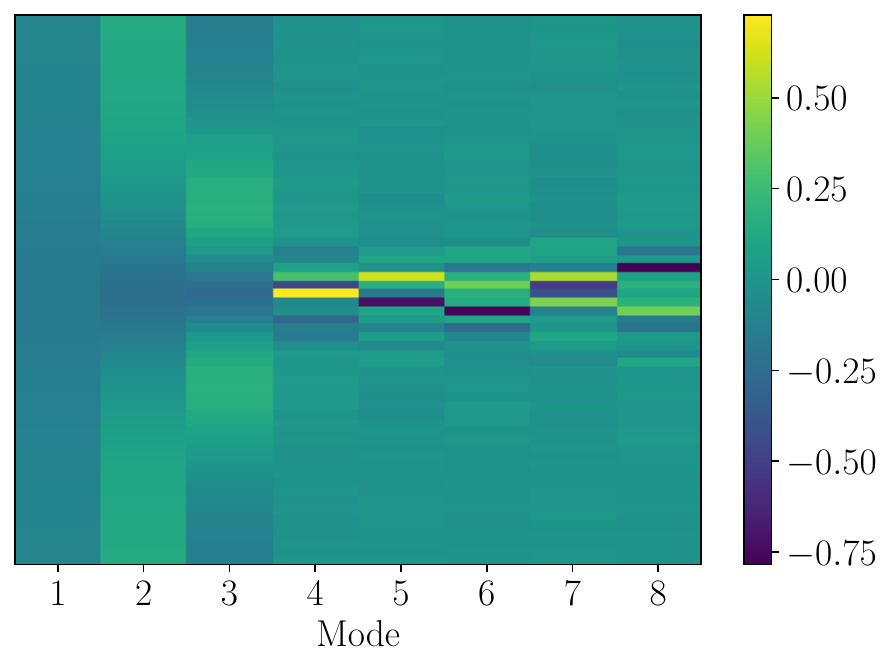}
        \caption{NLSE: Mode shapes obtained via SVD of the extended snapshot data. The modes become much noisier at $\dimr>3$, which explains the slight increase in parameter estimation error after $\dimr=3$ in Fig.~\ref{fig:param_error_vs_r}.}  
        \label{fig:noisy_mode_shapes}
\end{figure}

Lastly, we study the parameter estimation accuracy of the proposed reduced-dimensional learning approach for different reduced dimensions. Since we have randomness in the measurement noise and the Adam optimizer, we draw 20 realizations of data and train on each for every reduced dimension value from $\dimr=2$ to $\dimr=8$. The optimization uses the same procedure described earlier in this section. Box and whisker plots of the squared parameter errors are shown in Fig.~\ref{fig:param_error_vs_r}. The error is naturally quite high for $\dimr=2$ but decreases and levels off at $\dimr=3$, which indicates that the first three modes based on the SVD of the extended snapshot matrix are relatively clean. For $r>3$, we observe modes with low values of signal-to-noise ratios which leads to a marginal increase in the estimation error. The shapes of these modes for a single noise realization are shown in Fig.~\ref{fig:noisy_mode_shapes}.

\subsubsection{Conclusions}
This example showed that Algorithm~\ref{alg:lowd_learning} was effective at estimating a parameter from a 64-dimensional system in a reduced 8-dimensional subspace with 20\% multiplicative measurement noise. Since the computational cost of the likelihood evaluation scales cubically with state dimension $\dimx$, reducing the state dimension by a factor of eight results in evaluation that is roughly 512 times cheaper compared to evaluation with the full state dimension. Furthermore, it was shown that the reduced dimension could be as small as $\dimr=3$ while still achieving comparable parameter estimation accuracy as $\dimr=8$, suggesting potential savings of up to 9,709 times. Additionally, the posterior FOM was shown to share important physical structure to the underlying system, as shown by its conservation of mass, momentum, and Hamiltonian over time.

\section{Conclusions}
\label{sec:conclusions}
In this work, we presented a structure-preserving Bayesian framework for learning deep neural network parameterizations of nonseparable Hamiltonian systems from noisy data. This framework uses Gaussian filtering to compute a likelihood based on a stochastic dynamics model that allows for the effects of model and measurement uncertainty to be accounted for differently. Unlike past works which only considered additive Gaussian noise models, we showed that the algorithm can be tailored to other noise models and provided a filter for multiplicative noise as an example. The numerical experiments for low-dimensional systems demonstrated that the proposed Bayesian framework is data-efficient and robust to noise in the data, whereas the standard machine learning approach breaks down in the presence of noisy data. Moreover, the Bayesian framework outperformed the NSSNN, a state-of-the-art machine learning approach, when training on data with multiplicative uniform noise, demonstrating that the Gaussian filtering approach is not overly restrictive for non-Gaussian measurement noise.

We also proposed a novel algorithm for the identification of high-dimensional Hamiltonians that allows for cost-efficient parameter estimation through filtering in a low-dimensional symplectic subspace. Using prior knowledge about the underlying physics, this algorithm was effective in parameter estimation of a nonlinear Schr\"{o}dinger equation within $2.185\times10^{-3}$ mean squared error, even with data corrupted by 20\% multiplicative uniform noise. The full-order models based on this parameter estimate provided accurate and stable predictions, while also preserving the system Hamiltonian and other invariants of motion. Future work on this topic will explore estimation using partial observations, including improving low-dimensional projections derived from partially unknown/uncertain full-order models.

\section*{CRediT authorship contribution statement}

\textbf{Nicholas Galioto:} Conceptualization, Methodology, Investigation, Software, Formal analysis, Writing - original draft, Visualization. \textbf{Harsh Sharma:} Conceptualization, Methodology, Software, Writing - review \& editing. \textbf{Boris Kramer:} Conceptualization, Validation, Resources, Writing: Review \& Editing, Funding Acquisition. \textbf{Alex Arkady Gorodetsky:} Conceptualization, Methodology, Resources, Writing: Review \& Editing, Funding Acquisition, Supervision

\section*{Declaration of competing interest}
Alex Gorodetsky reports a relationship with Geminus AI that includes: employment and equity or stocks.
Boris Kramer reports a relationship with ASML Holding US that includes: consulting or advisory.
The other authors declare that they have no known competing financial interests or personal relationships that could have appeared to influence the work reported in this paper.

\section*{Data availability}
Data will be made available upon request.

\section*{Acknowledgments}
A.A. Gorodetsky and N. Galioto were funded by the AFOSR Computational Mathematics Program (P.M. Fariba Fahroo) under award FA9550-19-1-0013. B. Kramer and H. Sharma were in part financially supported by the Ministry of Trade, Industry and Energy (MOTIE) and the Korea Institute for Advancement of Technology (KIAT) through the International Cooperative R\&D program (No.~P0019804, Digital twin based intelligent unmanned facility inspection solutions) and the Applied and Computational Analysis Program of the Office of Naval Research under award N000142212624.

\appendix
\section{Cotangent lift algorithm}\label{sec:cotangent}
In projection-based model reduction, the semi-discrete model is projected onto a low-dimensional subspace. The key idea in structure-preserving model reduction is to preserve the underlying geometric structure during the projection. Since the FOM is a Hamiltonian system with underlying symplectic structure, the projection step is treated as the symplectic inverse of a symplectic lift from the low-dimensional subspace to the state space, see~\cite{peng2016symplectic}. A \textit{symplectic lift} is defined by $\y=\V\tilde{\y}$ where $\V \in \Real^{2d \times 2r}$ is a \textit{symplectic matrix}, i.e., a matrix that satisfies 
\begin{equation}
\V^\top\Jn\V=\Jr.
\end{equation}
The \textit{symplectic inverse} $\Vt\in \Real^{2r \times 2d}$ of a symplectic matrix $\V$ is defined by 
\begin{equation}
\Vt=\Jr^\top\V^\top\Jn,
\end{equation}
and the \textit{symplectic projection} can be written as $\tilde{\x}=\Vt\x\in \Real^{2r}$. Proper symplectic decomposition (PSD)~\cite{peng2016symplectic} is a method to find a symplectic projection matrix $\V$ that simultaneously minimizes the projection error in a least-squares sense, i.e., 
\begin{equation}
\min_{\substack{\V \\ \text{s.t.} \ \V^\top\Jn\V=\Jr}} \Vert \textbf{X} - \V\Vt \textbf{X} \Vert_F,
\label{eq:PSD}
\end{equation}
where $\textbf{X}:=[\x(t_1), \cdots, \x(t_K)] \in \Real^{2d \times K}$ is the snapshot data matrix, and $\Vert \cdot \Vert_F$ is the Frobenius norm. Since solving Eq.~\eqref{eq:PSD} to obtain the symplectic basis matrix $\V$ is computationally expensive, the authors in \cite{peng2016symplectic} outlined three efficient algorithms for finding approximated optimal solution for the symplectic matrix $\V$. These algorithms (cotangent lift, complex SVD, and a nonlinear programming approach) search for a near-optimal solution over different subsets of $Sp(2r,\Real^{2d})$, the set of all $2d \times 2r$ symplectic matrices. The cotangent lift algorithm computes the SVD of the extended snapshot matrix $\textbf{X}_e=[\q(t_1), \cdots,\q(t_K), \p(t_1), \cdots,\p(t_K)] \in \Real^{d \times 2K}$  to obtain a POD basis matrix $\mathbf{\Phi} \in \Real^{d \times r}$ and then constructs the symplectic basis matrix $\V= \begin{bmatrix}
      \mathbf{\Phi} & \bzero \\
     \bzero & \mathbf{\Phi}
     \end{bmatrix} \in \Real^{2d \times 2r}$ with $\Vt=\V^\top$. Compared to the complex SVD and the nonlinear programming approach, the cotangent lift algorithm is more easily implemented in the offline stage as it only requires the SVD of the extended snapshot matrix $\textbf{X}_e$. Moreover, the diagonal nature of $\V$ ensures that the interpretability of $\q$ and $\p$ is retained in the reduced setting.

\section{Hamiltonian operator inference}\label{sec:hopinf}
In this section, we describe how to estimate the terms $\Dhatq$ and $\Dhatp$ in the quadratic portion of the Hamiltonian~\eqref{eq:Hr_params} using the approach of H-OpInf~\cite{sharma2022hamiltonian}. First, assume we have an initial estimate of $\vtheta_{\text{nl}}$ defining the nonlinear terms $H_{\text{nl}}$. Then, define the nonlinear forcings $\fq$ and $\fp$ as 
\begin{align}
\fq(\x,\vtheta_{\text{nl}})&=\begin{bmatrix}\frac{\partial H_{\text{nl}}}{\partial p^1} (q^1,p^1,\vtheta_{\text{nl}})  \cdots \frac{\partial H_{\text{nl}}}{\partial p^{\nz}} (q^{\nz},p^{\nz},\vtheta_{\text{nl}})\end{bmatrix}^\top \in \reals^{\nz}, \\
\fp(\x,\vtheta_{\text{nl}})&=\begin{bmatrix}\frac{\partial H_{\text{nl}}}{\partial q^1} (q^1,p^1,\vtheta_{\text{nl}})  \cdots \frac{\partial H_{\text{nl}}}{\partial q^{\nz}} (q^{\nz},p^{\nz},\vtheta_{\text{nl}})\end{bmatrix}^\top \in \reals^{\nz}.
\end{align}
Utilizing the explicit forms of $\fq$ and $\fp$, form the nonlinear forcing snapshot matrices
\begin{equation}
\label{eq:F}
\Fq(\mQ,\mP,\vtheta_{\text{nl}})=
\begin{bmatrix}
      \fq(\x_1,\vtheta_{\text{nl}})  \cdots   \fq(\x_{\numsim},\vtheta_{\text{nl}})
\end{bmatrix}
, \qquad
\Fp(\mQ,\mP,\vtheta_{\text{nl}})=
\begin{bmatrix}
    \fp(\x_1,\vtheta_{\text{nl}}) \cdots   \fp(\x_{\numsim},\vtheta_{\text{nl}})
\end{bmatrix}.
\end{equation}
Next, project the trajectory~\eqref{eq:snapshot} and nonlinear~\eqref{eq:F} snapshot matrices onto the symplectic subspace using the projection matrix~\eqref{eq:cotangent} found with the cotangent lift~\cite{peng2016symplectic}
\begin{equation}
  \Qhat = {\mPhi}^\top\mQ \in \reals^{\dimr \times \numsim}, \qquad 
  \Phat = {\mPhi}^\top\Pp \in \reals^{\dimr \times \numsim}, \qquad
  \Fhatq = {\mPhi}^\top\Fq \in \reals^{\dimr \times \numsim}, \qquad 
  \Fhatp = {\mPhi}^\top\Fp \in \reals^{\dimr \times \numsim}.
\end{equation}
Additionally, compute the reduced time-derivative data $\dot{\qhat}$ and $\dot{\phat}$ from the reduced state trajectory data $\Qhat$ and $\Phat$ using a finite difference scheme. Then, organize the time-derivatives into snapshot matrices
\begin{equation*}
\dot{\Qhat}= \begin{bmatrix}
       \dot{\qhat}_1  \cdots \dot{\qhat}_{\numsim}
     \end{bmatrix}  \in \reals^{\dimr \times \numsim}, \quad \dot{\Phat}=\begin{bmatrix}
       \dot{\phat}_1  \cdots \dot{\phat}_{\numsim}
     \end{bmatrix}  \in \reals^{\dimr \times \numsim}.
     \label{eq:dQhat}
\end{equation*}
Lastly, infer the reduced operators $\Dhatq(\vtheta_{\text{quad}})$ and $\Dhatp(\vtheta_{\text{quad}})$ by solving the following constrained operator inference problem
\begin{equation}
\label{eq:H-OpInf_opt}
\min_{\substack{\Dhatq=\Dhatq^\top,\\ \Dhatp=\Dhatp^\top}} \bigg \lVert \begin{bmatrix}
\dot{\Qhat}- \Fhatq(\Qhat,\Phat,\vtheta_{\text{nl}}) \\ \dot{\Phat} + \Fhatp(\Qhat,\Phat,\vtheta_{\text{nl}}) 
\end{bmatrix} -  
\begin{bmatrix} 
\bzero & \Dhatp \\ 
-\Dhatq & \bzero
\end{bmatrix}
\begin{bmatrix}
 \Qhat\\ \Phat
\end{bmatrix} 
\bigg\rVert _F,
\end{equation}
where $\lVert\cdot\rVert_F$ denotes the Frobenius norm. The symmetric constraints on the reduced operators $\Dhatq$ and $\Dhatp$ ensure that the ROMs learned via H-OpInf are Hamiltonian. By introducing the terms $\Rhatp=\dot{\Qhat}- \Fhatq(\Qhat,\Phat)$ and $\Rhatq=-\dot{\Phat}-\Fhatp(\Qhat,\Phat)$, the above inference problem~\eqref{eq:H-OpInf_opt} can be reformulated as the Lyapunov equations
\begin{equation}
  \label{eq:lyap}
  (\Qhat\Qhat^\top)\Dhatq + \Dhatq(\Qhat\Qhat^\top)=\Qhat\Rhatq^\top + \Rhatq\Qhat^\top,
  \qquad
  (\Phat\Phat^\top)\Dhatp + \Dhatp(\Phat\Phat^\top)=\Phat\Rhatp^\top + \Rhatp\Phat^\top,
\end{equation}
which can be solved with off-the-shelf Lyapunov solvers.

The constrained operator inference problem in Eq.~\eqref{eq:H-OpInf_opt} enforces structure preservation at the reduced level to learn nonintrusive Hamiltonian ROMs that conserve the reduced Hamiltonian function $\Hr(\vqr,\vpr)$~\eqref{eq:Hr_params}. The authors in~\cite{sharma2022hamiltonian} have shown that the reduced Hamiltonian function $\Hr(\vqr,\vpr)$ can be interpreted as a perturbation of the FOM Hamiltonian $H$, i.e. $\Hr(\vqr,\vpr)=H(\Phi\qhat,\Phi\qhat) + \Delta H(\qhat,\phat)$. Consequently, the H-OpInf ROM yields approximate FOM trajectories, i.e. $\q_{\text{approx}}=\Phi\qhat$ and  $\p_{\text{approx}}=\Phi\phat$, that track the FOM solution trajectories accurately while also conserving a perturbed FOM Hamiltonian which yields bounded FOM energy error.

\bibliographystyle{elsarticle-num}
\bibliography{bib}

\end{document}